\pdfoutput=1

\documentclass[11pt]{article}

\usepackage{acl}

\usepackage{times}
\usepackage{latexsym}

\usepackage[T1]{fontenc}

\usepackage[utf8]{inputenc}

\usepackage{microtype}

\usepackage{amsmath}
\usepackage{amssymb}
\usepackage{multirow}
\usepackage{graphicx}
\usepackage{bm}

\DeclareMathOperator*{\argmax}{arg\,max}

\usepackage{svg}
\usepackage{relsize}
\usepackage{bbm}
\usepackage{array}

\usepackage[utf8]{inputenc} 
\usepackage[T1]{fontenc}    
\usepackage{hyperref}       
\usepackage{url}            
\usepackage{booktabs}       
\usepackage{amsfonts}       
\usepackage{nicefrac}       
\usepackage{microtype}      
\usepackage{xcolor}         
\usepackage{algorithm}
\usepackage{algpseudocode}
\usepackage{float}
\usepackage{comment}
\usepackage{graphicx}
\usepackage{svg}
\usepackage{multirow}
\usepackage{subcaption}
\usepackage{paralist}
\usepackage{enumitem}

\usepackage[backgroundcolor=yellow!30,linecolor=red, bordercolor=red,textsize=scriptsize]{todonotes}
\newcommand{\sys}{{\itshape DKAF}}
\newcommand{\incbabi}{{\textit{inc}-bAbI}}
\newcommand{\incBiTOD}{{\textit{inc}-BiTOD}}

\title{DKAF: KB Arbitration for Learning Task-Oriented Dialog Systems\\ with Dialog-KB Inconsistencies}

\author{Saley Vishal Vivek $^{1}$,  Rocktim Jyoti Das $^{1}$, Dinesh Raghu $^{1 \hspace{0.1cm} 2}$ and Mausam $^{1}$ \\
        $^{1}$ Indian Institute of Technology, New Delhi, India\\
        $^{2}$ IBM Research, New Delhi, India\\
        Vishal.Vivek.Saley@cse.iitd.ac.in, rocktimjyotidas@gmail.com \\ diraghu1@in.ibm.com, mausam@cse.iitd.ac.in}

\begin{document}
\maketitle
\begin{abstract}

Task-oriented dialog (TOD) agents often ground their responses on external knowledge bases (KBs). These KBs can be dynamic and may be updated frequently. Existing approaches for learning TOD agents assume the KB snapshot contemporary to each individual dialog is available during training. However, in real-world scenarios, only the latest KB snapshot is available during training and as a result, the train dialogs may contain facts conflicting with the latest KB. These dialog-KB inconsistencies in the training data may potentially confuse the TOD agent learning algorithm.


In this work, we define the novel problem of learning a TOD agent with dialog-KB inconsistencies in the training data. We propose a \textbf{D}ialog-\textbf{K}B \textbf{A}rbitration \textbf{F}ramework (\sys) which reduces the dialog-KB inconsistencies by predicting the contemporary KB snapshot for each train dialog. These predicted KB snapshots are then used for training downstream TOD agents. As there are no existing datasets with dialog-KB inconsistencies, we systematically introduce inconsistencies in two publicly available dialog datasets. We show that TOD agents trained with \sys{} perform better than existing baselines on both these datasets.

\end{abstract}

\section{Introduction}
\begin{figure*}
    \centering
    \includegraphics[scale=0.16]{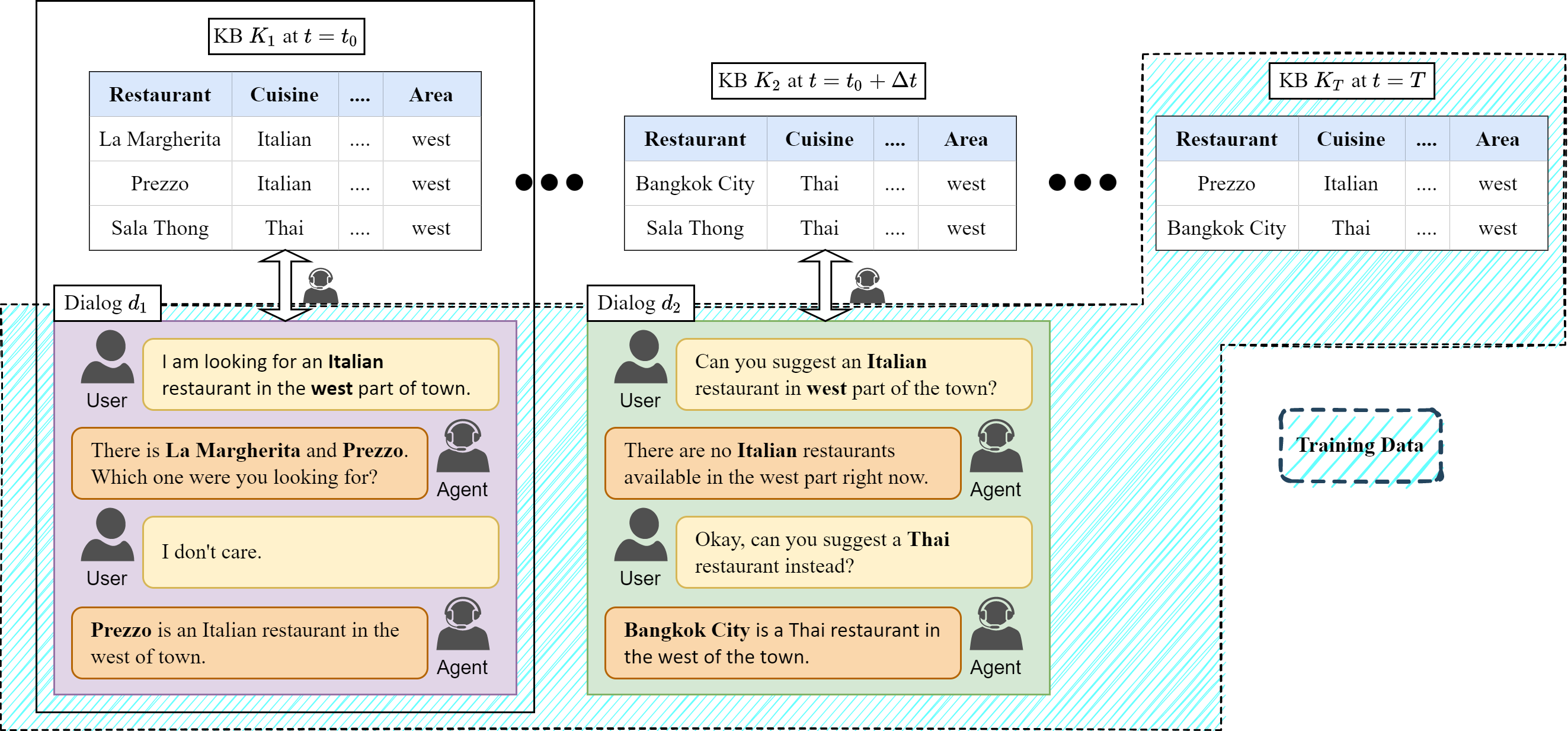}
    \caption{Figure shows snapshots of an evolving KB at times $t_0$, $t_0 + \Delta t$ and $T$. Over time, restaurants in the KB is changing, which is reflected in the KB snapshots $K_1$ and $K_2$ at time $t_0$ and $t_0 + \Delta t$ respectively. Dialogs $d_1$ and $d_2$ are consistent with KB snapshots $K_1$ and $K_2$. During training, KB snapshot $K_T$ is associated with dialogs $d_1$ and $d_2$ resulting in dialog KB inconsistencies. Shaded region defines our problem setting.}\label{fig:motivation}
\end{figure*}

A task-oriented dialog (TOD) system often requires information from a knowledge base (KB) to complete user goals like restaurant reservations, flight bookings, and calendar enquiry. This paper follows the recent line of research in \emph{end-to-end} approaches \cite{Wu2019GlobaltolocalMP, Qin2020DynamicFN, Raghu2021ConstraintBK}, where dialog agents are trained using just the training dialogs and an associated KB, without any expensive dialog state annotation. 

The KB contents typically change to reflect the transactions that happened during the user-agent dialogs. For example, in Figure \ref{fig:motivation}, the KB snapshot $K_1$ can transform into $K_2$ when \textit{La Margherita} and \textit{Prezzo} become unavailable due to reservations, and \textit{Bangkok City} becomes available due to a cancellation. Due to this evolving nature of the KB, two dialogs which started with the same user goal can result in two different outcomes. For example, consider the dialogs $d_1$ and $d_2$ in Figure \ref{fig:motivation}. In $d_1$, the agent makes two recommendations from $K_1$, whereas, in $d_2$, no recommendation is feasible as $K_2$ has no restaurants that fit the user's need.


Existing approaches for learning TOD agents assume the KB snapshot contemporary to each dialog is available during training. Such an assumption is limiting due to two reasons. First, KB snapshots are usually created at periodic intervals not after each KB transaction due to storage constraints. Second, dialogs used for training TOD models are often collected from messaging applications where human agents and users interact. Human agents often access the associated KB using a different application and so the KB queries fired during the dialog do not get logged with the dialogs \cite{Raghu2021UnsupervisedLO}. Without these KB query logs, it is difficult to reconstruct the contemporary KB.


As the contemporary KB snapshots are unavailable, a single KB snapshot (generally, the latest) is made available during training. When the latest KB snapshot gets associated with the train dialogs, the dialogs and the KB  may portray diverging information resulting in \textit{dialog-KB inconsistencies}. In the running example, $K_T$ denotes the latest KB snapshot. Dialog $d_1$ disagrees with $K_T$, as \textit{La Margherita} is missing from $K_T$. Dialog $d_2$ also disagrees with $K_T$, since $K_T$ contains an Italian restaurant, contradicting agent response.

Dialog-KB inconsistencies hinder the learning of TOD agents. These inconsistencies can force the TOD agent to either learn spurious patterns (e.g., using $d_2$ and $K_T$ may force the agent to ignore \textit{Prezzo}) or memorizes responses (using $d_1$ and $K_T$, will force the agent to generate \textit{La Margherita}) leading to poor generalization. To overcome these challenges, we define the novel problem of end-to-end learning of TOD systems with dialog-KB inconsistencies in training data. We also propose \sys{}, whose goal is to reduce the dialog-KB inconsistencies by predicting the contemporary KB for each dialog in the training corpus. These predicted KB snapshots and the associated dialogs can then be used to train any existing end-to-end TOD learning approaches.


Given a dialog, inconsistencies can be removed by inserting a new row in the KB based on the entities and relationships present in the dialog (e.g., adding \textit{La Margherita} to $K_T$ can make $d_1$ consistent with $K_T$). Inconsistencies can also be removed by deleting rows (e.g., removing \textit{Prezzo} from $K_T$ can make $d_2$ consistent). 
As dialogs offer \textit{weak supervision} to reduce dialog-KB inconsistencies, we use distant supervision and reinforcement learning to train \sys{}.

We construct two datasets by systematically infusing dialog-KB inconsistencies on bAbI \cite{Bordes2017LearningEG}, and BiTOD (English) \cite{Lin2021BiToDAB} datasets and refer to them as \incbabi{} and \incBiTOD{} respectively. 
Our experiments show that \sys{} reduces the dialog-KB inconsistencies and the overall TOD system trained with the KB predicted by \sys{} outperforms existing state-of-the-art models on both the datasets. In summary, 

\begin{enumerate}
\vspace{-1ex}
    \item We introduce the novel problem of training task-oriented dialog systems over data with dialog-KB inconsistencies.
\vspace{-1ex}
    \item We present \sys{} that alleviates dialog-KB inconsistencies by predicting the contemporary KB based on a given training dialog.
\vspace{-1ex}
    \item We systematically modify two publicly available datasets for the proposed task. Our experiments demonstrate that \sys{} improves TOD performance on these datasets. 
\end{enumerate}
We release all resources for future research\footnote{\href{https://github.com/dair-iitd/DKAF}{https://github.com/dair-iitd/DKAF}}.

\section{Related Work}
Traditionally, dialog systems are modular \citep{Young2013POMDPBasedSS, RojasBarahona2016ANE, HosseiniAsl2020ASL} with different modules for natural language understanding, dialog state tracking, and natural language generation. These models require hand-crafting of dialog states and require expensive intermediate annotations for training each component. On the other hand, end-to-end TOD models \citep{Eric2017KeyValueRN, Madotto2018Mem2SeqEI, Raghu2021ConstraintBK, Raghu2019Disentangle, Wu2019GlobaltolocalMP} that directly predict system response given dialog history and the KB are becoming increasingly popular as they alleviate the need for expensive annotations.
\sys{} approach proposed in this work focuses on learning end-to-end TOD system when training data has dialog-KB inconsistencies.

Recent works on inconsistency in dialog generation by \citet{Nie2021ILF, Qin2021DontBC, Qin2020DynamicFN} study problem of detecting inconsistent dialog responses with respect to dialog history, user intent, the KB. \citet{Welleck2019DialogueNL} explores a similar problem but in domain of Persona-based dialog systems. \citet{Larson2020InconsistenciesIC} studies the topology of annotation inconsistencies in crowd-sourced data for slot-filling models.

\sys{} differs from these works in two key ways: (1) its objective is learning a TOD model when training data includes dialogs inconsistent with the KB and, (2) it explicitly resolves dialog-KB inconsistencies via a novel KB arbitration procedure.

\section{Problem Definition}
We first describe the task of learning an end-to-end TOD system. We denote a dialog between user $u$ and agent $a$ as $d = [u^u_1, u^a_1, u^u_2, u^a_2,...,u^u_m, u^a_m]$ where $m$ denotes number of exchanges. Let $\{d_j\}_{j=1}^{N}$ be the set of $N$ training dialogs. An end-to-end TOD system predicts agent response $\hat{u}^a_i$ given dialog history $[u^u_1, u^a_1, u^u_2, u^a_2,...u^u_i]$ and an associated KB $K_T$. This system is trained using $\{d_j, K_T\}_{j=1}^{N}$ where $K_T$ is assumed to be consistent with all the training dialogs.

We now consider the setting where training dialogs are grounded in an evolving KB. Here, a training dialog $d_j$ is consistent with its contemporary KB snapshot, $K_j$. However, at training time, a single KB snapshot $K_T$ is available which gets associated with all training dialogs resulting in dialog-KB inconsistencies. So, we propose the task of learning end-to-end TOD system using $\{d_j, K_T\}_{j=1}^{N}$ with dialog-KB inconsistencies.

\section{\sys}
\begin{figure*}
    \centering
    \includegraphics[scale=0.12]{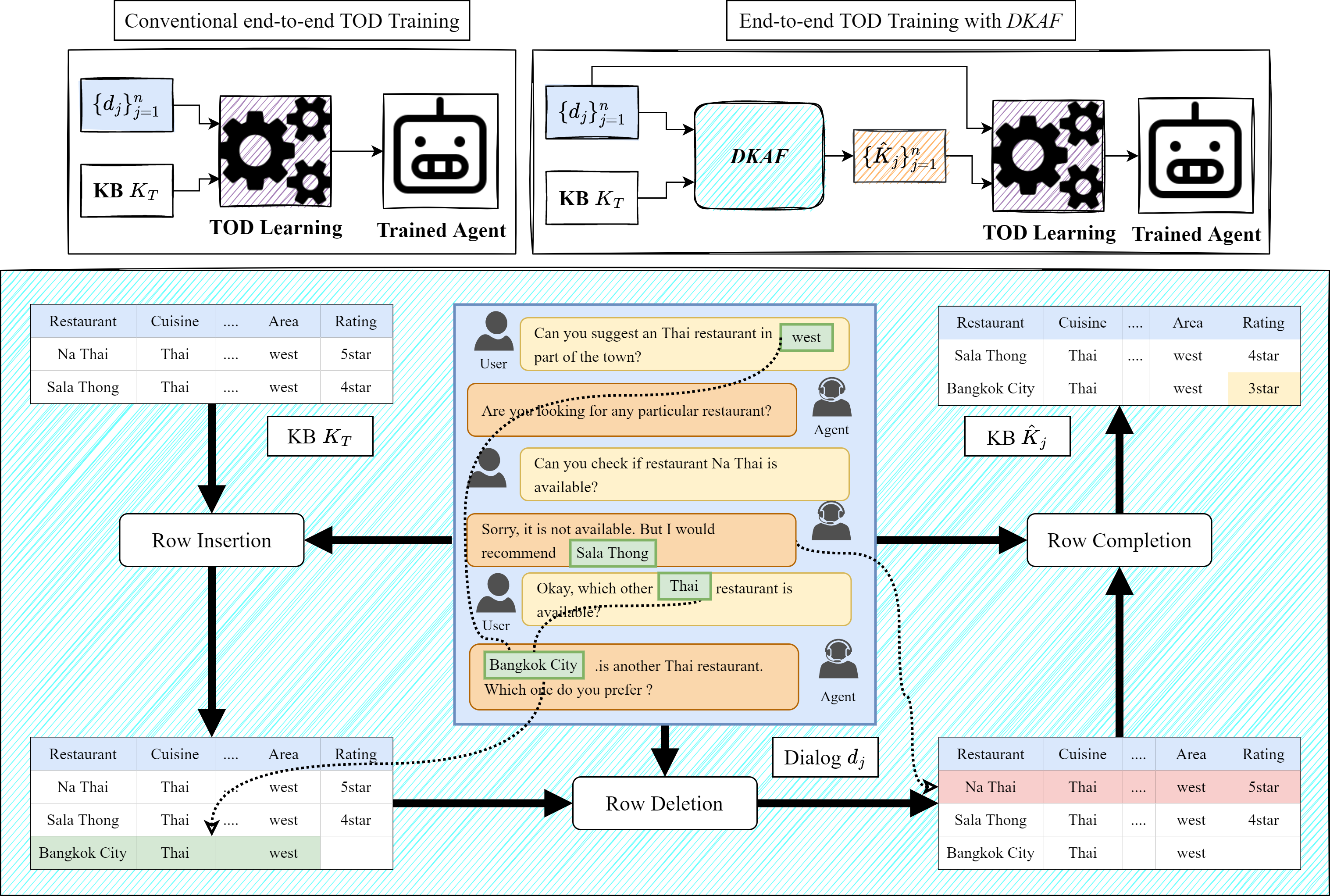}
    \caption{Comparison of conventional TOD learning (top-left) with TOD learning with \sys{} (top-right). \sys{} attempts to resolve dialog-KB inconsistencies by updating training KB $K_T$ given a training dialog. Figure (bottom) shows \sys{} in action with KB updates from row insertion, row deletion and row completion to training KB $K_T$.}
    \label{fig:pipeline}
\end{figure*}

To solve dialog-KB inconsistencies, we propose \sys{} that updates $K_T$ based on $d_j$ such that the resultant KB snapshot $\hat{K}_j$ resembles with $K_j$. A TOD system is then trained using $\{d_j, \hat{K}_j\}_{j=1}^N$.
\sys{}'s updates to $K_T$ happen through a cascade of three models - row insertion, row deletion, and row completion. Each model takes the KBs resulting from the preceding model and performs modifications to them based on the training dialogs. Figure \ref{fig:pipeline} highlights this process. We now describe each model in detail.

\subsection{Row Insertion (RI)}\label{sec:ri}
Row insertion aims to extract rows from the dialogs that are missing from the training KB. For this, RI model predicts if a relation $r$ holds between entities $e_1$ and $e_2$ mentioned in a given dialog $d$. Following \citet{Zhang2015RelationCV}, it infuses $d$ with position indicators for $e_1$ and $e_2$ and encodes the resulting dialog using a hierarchical encoder \citep{Sordoni2015AHR}.
Encoder feature vectors for a dialog and entities are then passed through classifier network for relation $r$. Thus, RI model uses training dialog to identify missing KB relationships $(e_1, r, e_2)$. Figure \ref{fig:pipeline} showcases this where \textit{(Bangkok City, cuisine, Thai)} and \textit{(Bangkok City, area, west)} get added to the KB. We provide more details in \ref{app:ri_arch}.

We form supervised data for training RI model with distant supervision and follow annotation scheme of \citet{Xu2013FillingKB}. Given a training dialog $d$, we form three sets - positive, negative and infer consisting of type-consistent relationships.
For entities $e1, e2 \in d$\footnote{can be identified by NER, though in this work, we assume this is known}, a relationship $(e_1, r, e_2)$ is in positive set if it also exists in $K_T$. 
A relationship $(e_1, r, e_2)$ is in negative set when its head entity $e_1$ exists in $K_T$ but the relationship does not. We follow this conservative annotation to avoid to false negatives samples. We add all remaining relationships to infer set.
We train RI model over the union of positive and negative sets from all training dialogs.

We apply RI model over infer set from training dialog $d_j$ to obtain KB snapshot $K^{ri}_{j}$ post insertion.

We note that \citep{Yu2020DialogueBasedRE} proposed a similar task of predicting relations among the individuals engaged and mentioned in dialogs from a popular TV series. However their approach is fully supervised while we use distant supervision.

\subsection{Row Deletion (RD)}\label{sec:rd}
RD model predicts whether a row $\rho$ from KB $K$ (mis)aligns with a given dialog $d$. Here, $\rho$ is misaligned if it disrupts agent reasoning in $d$. In figure \ref{fig:pipeline}, row for \textit{Na Thai} is misaligned with $d_j$ since it forces the TOD system to generate a factually incorrect response \textit{"Sorry it is not available..."}.
Further, it hinders TOD system from producing \textit{Sala Thong} as it is rated below \textit{Na Thai}. We use RD model predictions to drop misaligned rows from the KB.

For input $d$, RD model computes dialog features using the dialog encoder given in Section \ref{sec:ri}. Recent works \citep{Banerjee2019GraphCN,Yang2020GraphDialogIG} showcase the efficacy of GCNs in TOD modeling. Consequently, RD model includes an r-GCN \citep{Schlichtkrull2018ModelingRD} KB encoder that computes KB entity features. Then, RD model reasons over KB entities using a memory network \citep{Sukhbaatar2015EndToEndMN} with dialog features as query input.
Finally, it appends memory network output with features of a row (sum of constituent entity features). The resulting vector is fed to a feed-forward network that makes binary prediction. We provide further information in \ref{app:rd_arch}

\subsubsection*{Training RD Model} 
We adopt reinforcement learning (RL) to train RD model due to lack of supervised dataset. We treat RD model as an RL agent that inputs a state $(d, K, \rho)$ and takes action $a\in \{0, 1\}$ where $a=0$ means $\rho$ is misaligned with $d$.
Given reward function $R_a(d, K, \rho)$, RL objective for training RD is
\begin{equation*}
    J_{RD} = \sum_{j=i}^N \frac{1}{|K^{ri}_j|} \sum_{\rho \in K^{ri}_j} R_a(d_j, K^{ri}_j, \rho)
\end{equation*}
We posit that a TOD system can provide an appropriate reward function for the task. In our running example, dropping \textit{Na Thai} from the KB aids agent reasoning in the dialog causing likelihood of \textit{Sala Thong} in the agent utterance to improve. Thus, likelihood score from a TOD system can guide RD tasks.
We incorporate this insight using a novel masked entity modeling (MEM) task. Let $e$ be an entity in the $i^{th}$ utterance in given dialog $d$. We form a masked dialog history $H_e$ consisting of utterances till $i^{th}$ utterance and replace entity $e$ in $i^{th}$ utterance with a \textit{<mask>} token. Let $E_a$ be the set of entities occurring in agent utterances $d$. MEM objective is then to maximize following likelihood
\begin{equation}\label{eq:mem_fn}
    \mathcal{L}(d, K) = \prod_{e \in E_a} P(e|H_e, K)
\end{equation}
Now we derive reward function for RD model as
\begin{align*}
    R_0(d, K, \rho) &= sgn [\mathcal{L}(d, K \setminus \{\rho\}) - \mathcal{L}(d, K)] \\
    R_1(d, K, \rho) &= -R_0(d, K, \rho)
\end{align*}

Note that, deleting a conflicting row improves the likelihood in equation \ref{eq:mem_fn} thus incurs a positive reward otherwise a negative reward. 

Inspired by recent works \citep{Wu2019GlobaltolocalMP, Raghu2021ConstraintBK, He2020TaskOrientedDG}, we design our MEM model as a dual pointer network where $P(e|H_e, K)$ is modelled as probability of copying masked entity $e$ from $H_e$ tokens and KB entities. We discuss MEM model in detail in appendix \ref{app:mem_arch}.

We train both MEM and RD models using $\{d_j, K^{ri}_j\}_{j=1}^N$. We train RD using MAPO algorithm \cite{Liang2018MemoryAP}, since our action space is discrete and state transitions deterministic. We use predictions from RD model over $(d_j, K^{ri}_j, \rho)$ states from each $d_j$ to obtain snapshot $K^{rd}_j$ post deletion. 

\subsection{Row Completion (RC)}\label{sec:rc}
RI model adds new rows to the KB, which can be incomplete since fields like rating of restaurants need not occur explicitly in the dialog. Yet, these fields can be crucial for TOD system. Rating can be necessary, for example, when agent selects the restaurant from the KB based on its rating. 
We call fields like rating latent fields and RC model aims to deduce the values for such fields from the dialog. For example in figure \ref{fig:pipeline}, RI should predict a rating \textit{3star} or lower for \textit{Bangkok City}.

We consider entity $e_s$ in dialog $d$ such that $e_s$ is not related to any entity in KB $K$ via latent field type $r$. RC model aims to predict target entity for the partial relationship $(e_s, r)$ given $d$. It infuses $d$ with position indicators for $e_s$ and encodes resulting dialog using dialog encoder.
Similar to \ref{sec:rd}, it computes KB entity features using KB encoder and reasons over them using memory network.
Finally, it appends memory network output with $e_s$ encoding and feeds it to a feed-forward network that predicts the target entity $e_t \in E_r$. Here, $E_r$ is the set of valid target entities for $r$ based on the task ontology. We provide more details in \ref{app:ri_arch}


Similar to \ref{sec:rd}, we treat RC model as RL agent that observes state $(d, e_s, r, K)$ and takes an action $e_t \in E_r$. We use following reward function to train the model
\begin{equation*}
\begin{split}
    & R_{e_t}(d, e_s, r, K) = \\
    & \left\{
    	\begin{array}{ll}
    		1  & \mbox{if } e_t = \argmax_{e \in E_r} \mathcal{L}(d, K \cup \{e_s, r, e_t)\} ) \\
    		0  & \mbox{otherwise}
    	\end{array}
	\right.
\end{split}
\end{equation*}

For training dialog $d_j$, we create state space $\{(d_j, e_s, r, \Tilde{K}^{rd}_j)\}$ where entity $e_s \in d_j$, $r$ is a latent field and $\Tilde{K}^{rd}_j$ is formed by dropping any relationships $(e_s, r, e)$ from $K^{rd}_j$. We train RC model using MAPO over state-spaces combined over training dialogs. Finally, the trained RC model makes prediction over incomplete rows in $K^{rd}_j$ to get final snapshot $\hat{K}_j$.


\section{Experimental Setup}
\subsection{Datasets Construction}

Existing TOD datasets make a simplistic assumption that KB contents do not change over time. Hence, all dialogs in these datasets are consistent with the KB. To study our problem, we systematically induce dialog-KB inconsistencies in two existing TOD datasets, namely bAbI dialog \cite{Bordes2017LearningEG} \& BiTOD (English) \cite{Lin2021BiToDAB} and refer to them as \incbabi{} and \incBiTOD, respectively. 
bAbI dialog dataset consists of synthetically generated dialogs from the restaurant reservation domain. BiTOD is a human-generated multi-domain dialog dataset with dialogs in English and Chinese. For our experiments, we only use the English dialogs from hotel, restaurant, and attraction domains. For more details on these datasets please refer to Appendix \ref{sec:data-stats}.

\begin{figure*}[h]
    \centering
    \includegraphics[scale=0.12]{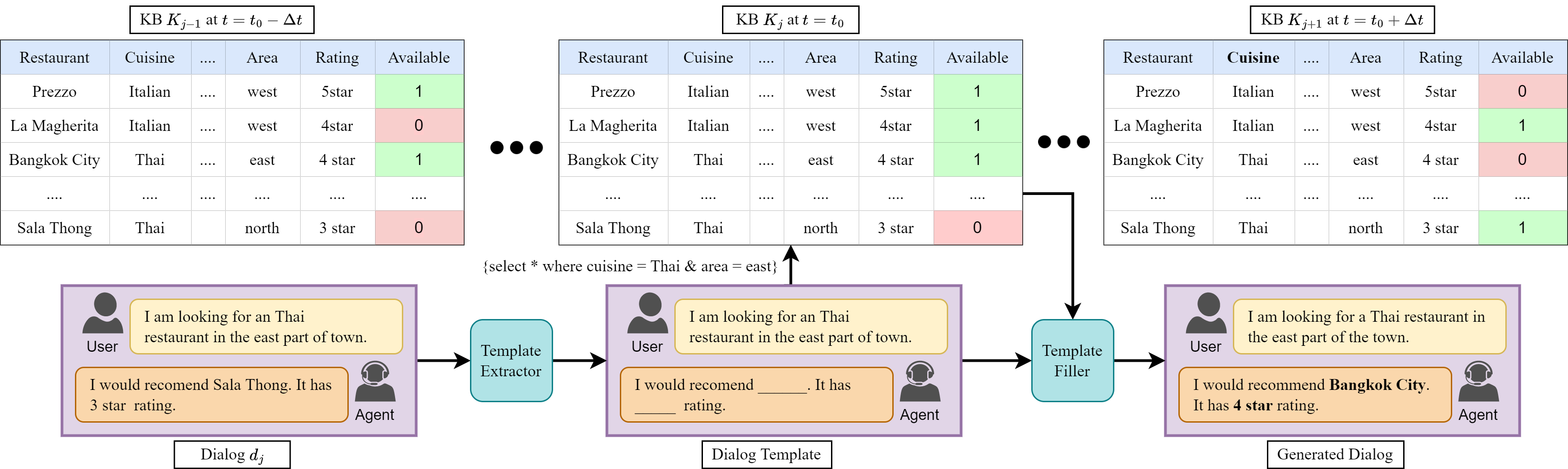}
    \caption{Figure shows the simulation pipeline used for generating datasets.}
    \label{fig:kb simulation pipeline}
\end{figure*}

We follow a two-step procedure to simulate the dialog-KB inconsistencies. In the first step, we generate an evolving KB by modifying its contents over time and maintaining a snapshot with timestamp associated with it. To generate an evolving KB, we add a binary random variable, named \textit{available}, to indicate the availability of each KB entry as illustrated in Figure \ref{fig:kb simulation pipeline}.

For restaurants, we wanted our simulator to reflect real-life scenarios where restaurants are often available during afternoons but are busy during peak hours (like evening and breakfast). To this end, we use the Yelp dataset\footnote{\url{https://www.yelp.com/dataset}}. Yelp provides the number of customers that have checked in into a restaurant at any given hour of the day for any day of the week. We use this data to simulate the availability of restaurants in our KB. Given the time of the day and day of the week, we sample restaurant availability to be inversely proportional to the number of check-ins from Yelp data. In our simulation, we also mimic (a) maintenance breaks by making restaurants unavailable for a day with a probability of 0.05 and (b) permanent closures with a probability of 1e-5.

Unfortunately, for hotels we did not find any check-ins data. we set the availability of each KB entry following a Bernoulli distribution parameterized by a success probability $p$ set to 0.75. Contrary to restaurants and hotels, attractions are generally available. Thus, we do not simulate their availability. Note that as entities are simulated differently, our dataset has a mixture of different evolving KB patterns.

In the second step, we assign a timestamp to each dialog and associate it with a corresponding KB snapshot. For example, the dialog $d_j$ in Figure \ref{fig:kb simulation pipeline} is associated with the snapshot $K_j$.  We then identify the KB entities present in the dialog (e.g., \textit{Sala Thong} and \textit{3 star} in $d_j$) and replace them with appropriate entities from the snapshot $K_j$ that match the annotated dialog state (e.g., \textit{cuisine=Thai, area=east)}. All modified dialogs and the last snapshot of the KB together form the inconsistent version of the dataset. Each modified dialog $d_j$ will be consistent with its KB snapshot $K_j$ but may not be consistent with the last snapshot used for training. To mimic real-world settings, we only induce inconsistencies in the train dialogs. The test dialogs remain consistent. 

\subsection{Algorithms}
We compare our proposed approach against the following baselines: GLMP \cite{Wu2019GlobaltolocalMP}, CDNet \cite{Raghu2019Disentangle} and SimpleTOD \cite{HosseiniAsl2020ASL}. GLMP and CDNet are both end-to-end TOD models. SimpleTOD is GPT2 based model that requires belief state annotations. So, we adapt SimpleTOD to the end-to-end TOD setting. For more details please refer to Appendix \ref{app:simpletod}.

We train the baselines on \incbabi{} and \incBiTOD{} datasets and identify the best-performing baseline. The best baseline is then trained in the following two settings:

\vspace{0.5ex}
\noindent \textbf{Rule-based:} A rule-based system performs KB arbitration for each dialog. Resulting KB snapshots are then used to train the TOD model. We defer the discussion of the rules in Appendix \ref{app:rule_based}.

\vspace{0.5ex}
\noindent \textbf{\sys{}:} This is our proposed approach that performs KB arbitration for each dialog $d_j$ with \sys{}. The predicted KB snapshot and dialog $\{d_j, \hat{K}_j\}_{j=1}^{N}$ pairs to train the TOD model.

\noindent The training details are reported in Appendix \ref{app:baseline_training}.

\subsection{Evaluation Metrics}

As \incbabi{} is synthetically generated, following \newcite{Bordes2017LearningEG}, we use exact string matching metrics:  response accuracy (percentage of predicted responses that exactly match the gold response) and dialog accuracy (percentage of dialogs with all correctly predicted responses).

As \incBiTOD{} is human-generated, we follow \newcite{Wu2019GlobaltolocalMP} and use BLEU \cite{papineni-etal-2002-bleu} and Entity F1 \cite{Eric2017KeyValueRN} for measuring response prediction performance. Dialog-KB inconsistencies can cause models to learn incorrect KB reasoning patterns. To measure this effect, we also report KB Entity F1 from \citet{Raghu2021UnsupervisedLO} computed for entities that can only be inferred from KB. We also perform human evaluation for \incBiTOD{} along two dimensions: (i) \textit{Relevance:} how useful are the responses given the dialog and KB, and (ii) \textit{Naturalness:} how human-like are the predicted responses. Each dimension is annotated on a Likert scale of 0-4 \cite{Likert1932ATF}.

\section{Results}
We answer the following research questions in our experiments:
\begin{compactenum}
\item \textit{Performance Study:} How effective is \sys{} in fixing the dialog-KB inconsistencies?
\item \textit{Ablation Study:} What is the performance gain from each component of \sys{}?
\item \textit{Incremental Analysis:} How robust is \sys{} to the number of inconsistent dialogs in the train data?
\end{compactenum}


\begin{table*}[h]
\centering
\small
\begin{tabular}{lccccc}
\toprule
\multirow{2}*{\textbf{Model}}  & \multicolumn{2}{c}{\textbf{\incbabi{}}} & \multicolumn{3}{c}{\textbf{\incBiTOD{}}} \\ 
\cmidrule(r){2-6}
& \textbf{Dialog Acc.}  &	\textbf{Response Acc.}    & \textbf{BLEU} & \textbf{Ent. F1}  & \textbf{KB Ent. F1}    \\
\midrule
GLMP                          & 73.6                     & 97.87                      & 15.29         & 0.674  & 0.633    \\
CDNet                         & 66.8                     & 96.76                      & 19.37         & 0.772  & 0.745    \\
SimpleTOD                     & 90.6                     & 99.39                      & 20.28         & 0.786  & 0.757    \\
\midrule
SimpleTOD + Rule-based        & 53.1                     & 96.28                      & 21            & 0.761  & 0.773    \\
SimpleTOD + DKAF              & \textbf{99.2}                     & \textbf{99.94}                      & \textbf{24.91}         & \textbf{0.819}  & \textbf{0.833}    \\
\bottomrule
\end{tabular}
\caption{Performance of GLMP, CDNet and SimpleTOD on \incbabi{} and \incBiTOD{} dataset. We report SimpleTOD in Rule-based and \sys{} setting. }
\label{tab:dkaf_results}
\end{table*}

\begin{table}[h]
\small
\centering
\begin{tabular}{lcc}
\toprule
 &	\textbf{Relevance} &		\textbf{Naturalness} \\
\midrule
SimpleTOD   & 	3.15 &	3.71 \\
SimpleTOD + Rule-based   & 	3.05 &	3.84 \\
SimpleTOD + DKAF & 	3.36 &	3.74 \\
\bottomrule
\end{tabular}
\caption{Human Evaluation on \incBiTOD{}}
\label{tab:human} 
\end{table}

\subsection{Performance Analysis}

Table \ref{tab:dkaf_results} reports the response prediction performance on \incbabi{} and \incBiTOD{} datasets. We first discuss the performance of baseline models. We then integrate \sys{} into the best-performing model - SimpleTOD and discuss how well \sys{} mitigates the effect of dialog-KB inconsistencies.


\vspace{0.5ex}
\noindent \textbf{Baseline Performance:}
We observe that dialog-KB inconsistencies affect baseline models in varying degrees. On \incbabi{} dataset, SimpleTOD achieves the best performance with 90.6\% dialog accuracy. Whereas, GLMP and CDNet perform poorly with dialog accuracy of 73.6\% and 66.8\%.

SimpleTOD also achieves the best performance on \incBiTOD{} dataset across all the metrics. This is expected, especially in the human-generated \incBiTOD{} dataset, as SimpleTOD is built on top of GPT2. We select SimpleTOD for our further experiments with \sys{}.

\vspace{0.5ex}
\noindent\textbf{Efficacy of \sys:}
We report the performance of SimpleTOD + Rule-based and SimpleTOD + \sys{} in table \ref{tab:dkaf_results}. In \incbabi{} dataset, SimpleTOD + \sys{} shows improvement over SimpleTOD model with 8.6\% gain in dialog accuracy. SimpleTOD is also the best-performing model across all baselines. To analyze the results of \sys{}, we compare the number of dialog-KB inconsistencies in \incbabi{} before and after \sys{} arbitration. \sys{} performs total of 239 insertions and 207 deletion in \incbabi{} causing inconsistencies to drop from 35.8\% to 2.8\% validating effectiveness of \sys{} in resolving the inconsistencies.

SimpleTOD + Rule-based, on contrary, performs worse even compared to SimpleTOD baseline. Rule-based arbitration performs 239 insertions and 1014 deletions to \incbabi{} reducing the inconsistency rate to 0\%. Yet, this does not result in performance improvement over baselines. Here, excessive deletions due to rule-based arbitration upset reasoning patterns in the dataset more than dialog-KB inconsistencies. Note that domain experts can improve such rule-based system further by incorporating reasoning patterns peculiar to the domain. On other hand, \sys{} makes achieves gains in performance with minimal domain-specific assumptions.

For \incBiTOD{} dataset, SimpleTOD + \sys{} outperforms SimpleTOD model in entity F1 and entity F1 KB metrics by a margin of 3.25 and 7.64 points. The gain in entity F1 KB is indicative of \sys{}'s effectiveness in resolving inconsistencies. In total, \sys{} makes 264 insertions and 207 deletions to \incBiTOD{} which results in dialog-KB inconsistencies to drop from 23\% to 6.94\%. We find that resolving dialog-KB inconsistencies is much more challenging in human-generated dataset. As in \incbabi{}, SimpleTOD + Rule-based under-performs compared to SimpleTOD baseline in \incBiTOD{} as well. Rule-based arbitration results in 5.08\% inconsistencies from 264 insertions and 2889 deletions.


\vspace{0.5ex}
\noindent \textbf{Human Evaluation:} We summarize the human evaluation results on the \incBiTOD{} dataset in Table \ref{tab:human}. We randomly sample 50 (dialog-context, response) pairs from \incBiTOD{} and two human judges labelled responses generated by SimpleTOD, SimpleTOD + Rule-based and SimpleTOD + \sys{} on relevance and grammar on a Likert scale (0-4) \cite{likert1932technique}. We observe that on relevance, SimpleTOD + \sys{} out-performs both SimpleTOD (0.21) and SimpleTOD + Rule-based (0.31) baselines.

However, naturalness score of SimpleTOD + Rule-based is better than SimpleTOD and SimpleTOD + \sys{}. Upon investigation, we found that the annotator favoured SimpleTOD+Rule-based due to minor grammatical errors. For example, the annotator preferred SimpleTOD+Rule-based because it used the preposition "from" instead of "on" before april 24 as shown below:
\begin{compactenum}
    \item \textit{SimpleTOD + Rule-based}: so you would like to book 4 rooms at mingdu hotel for 4 nights starting from april 24 ?
    \item  \textit{SimpleTOD + \sys{}}: so you would like to book 4 rooms at mingdu hotel for 4 nights starting on april 24 ?
\end{compactenum}
We provide more details on human evaluation in Appendix \ref{app:human_eval}. 

\subsection{Ablation Experiments}\label{exp:ablation}
\begin{table}[h]
\centering
\small
\resizebox{0.46\textwidth}{!}{
\begin{tabular}{lccc}
\toprule
\multirow{2}*{\textbf{Model}}& \multicolumn{1}{c}{\textbf{\incbabi{}}}& \multicolumn{1}{c}{\textbf{\incbabi{}(M)}}& \multicolumn{1}{c}{\textbf{\incBiTOD{}}}\\ 
\cmidrule(r){2-4} &	\textbf{Dlg Acc.}   & \textbf{Dlg Acc.}     &		\textbf{KB Ent. F1} \\
\midrule
SimpleTOD                                       & 90.6                     & 49.7                       & 0.757             \\
\midrule
 + DKAF w\textbackslash{}o RI                      & 91.9                     & 62.3                       & 0.749             \\
 + DKAF w\textbackslash{}o RD                      & 98                       & 77.7                       & 0.793             \\
 + DKAF w\textbackslash{}o RC                      & 99                       & 79.9                       & 0.833             \\
 + DKAF                                            & \textbf{99.1}                     & \textbf{88.6}                       & \textbf{0.833}             \\
\bottomrule
\end{tabular}}
\caption{Ablation Results}
\label{tab:ablations}
\end{table}

We perform ablation for each component in \sys{} to measure how each stage contributes to overall \sys{} performance. Table \ref{tab:ablations} reports our results.

For both \incbabi{} and \incBiTOD{}, excluding  RI leads to a significant performance drop. In the case of \incBiTOD{}, we observe that excluding RI also causes RD model to abstain from removing rows from the KB. Dropping RD results in performance drop of 1.1 points for \incbabi{} dataset and 0.04 for \incBiTOD{}. This is expected as agent suggestions in both \incbabi{}, and \incBiTOD{} follow rating orders, and row deletion restores this order by systematically deleting upsetting rows. This can be seen in examples given in table \ref{tab:babi_bitod} and \ref{tab:dkaf_bitod}. We provide further details on why dropping RI leads to severe degradation in comparison to RD and RC in section \ref{app:order}.


Finally, excluding RC has a lower impact in \incbabi{}. In \incbabi{}, restaurant names carry much of its attributes include its rating. We posit that SimpleTOD tokenization allows model a direct access to this rating. For example, SimpleTOD tokenizer splits restaurant name resto\_ rome\_cheap\_thai\_2stars in \incbabi{} into attributes (rest, o,\_ , rome, \_, che, ap, \_, th, ai, \_, 2, stars). As a result, SimpleTOD can operate sufficiently well even in absence of the ratings.

To validate this, we modify \incbabi{} dataset where we replace the rating in restaurant names with random alphabets. For example, we replace \textit{resto\_rome\_cheap\_thai\_2stars} with \textit{resto\_rome\_cheap\_thai\_Qstars}.  We report ablations on resulting dataset, named \incbabi{}(M), in table \ref{tab:ablations}. SimpleTOD performance significantly deteriorates in \incbabi{}(M) with a drop as high as 40.9 points compared to \incbabi{}. Note that \sys{} improved performance by a margin of 38.9 points. Here, we observe that excluding RC leads to 8.7 point drop. On the other hand, \incBiTOD{} does not have any such latent entities in the KB, thus resulting in no change in performance.


\subsection{Incremental Analysis}
\begin{figure}[h]
    \centering
    \includegraphics[scale=0.5]{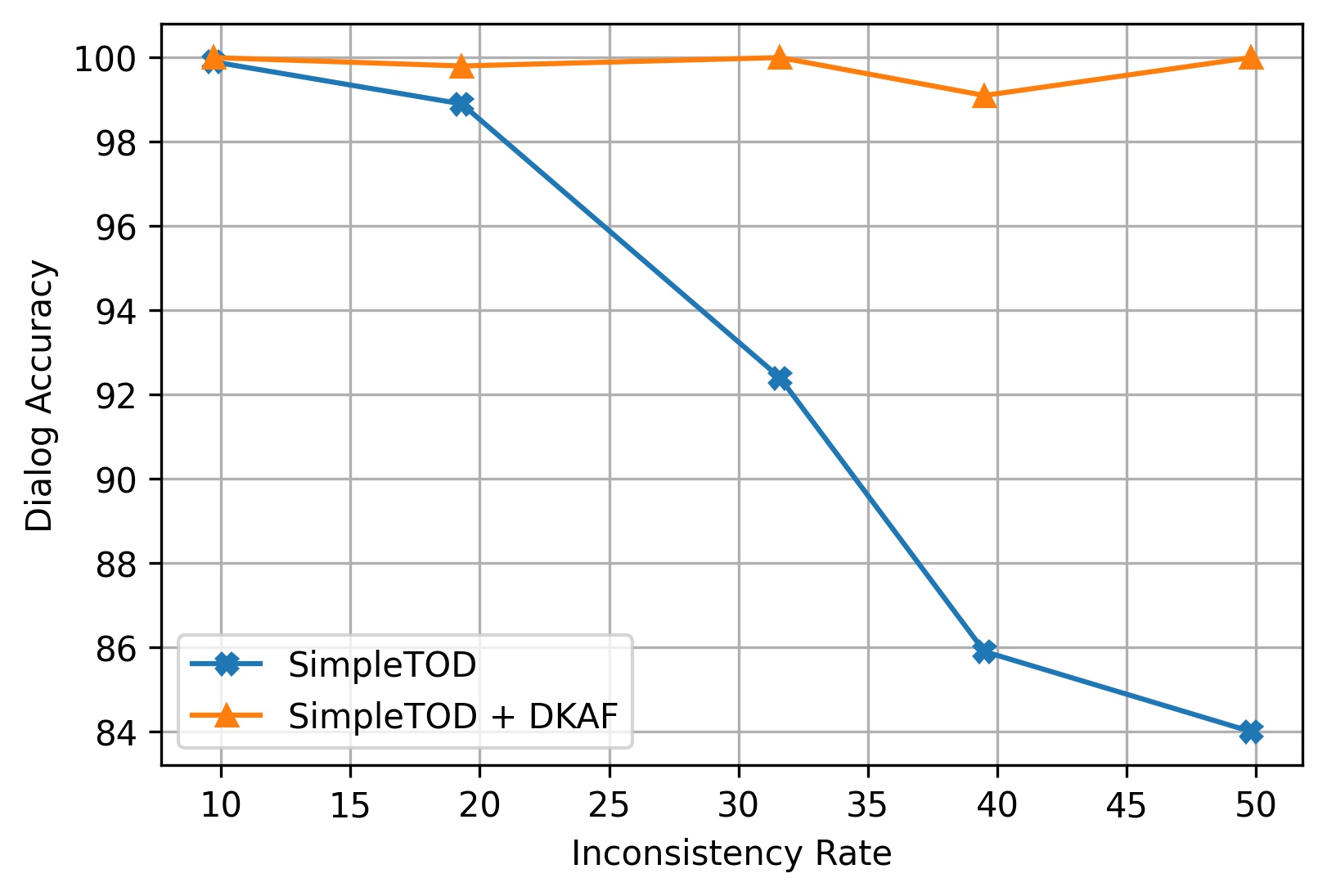}
    \caption{DKAF Incremental Analysis on \incbabi{}}
    \label{fig:babi_incr}
\end{figure}

We create 5 variants \incbabi{} dataset with increasing inconsistency rates in our simulation. For each dataset variant, we train SimpleTOD and SimpleTOD + \sys{} model. Figure \ref{fig:babi_incr} showcases the results. With an increasing number of dialog-KB inconsistencies, the performance of SimpleTOD model decreases sharply. On the other hand, SimpleTOD + \sys{} is consistently able to recover from the performance drop with significant gains. 

\subsection{Order of models in \sys{}}\label{app:order}
In this section, we validate our choice of order among the different models in \sys{}. As discussed in  section \ref{sec:rc}, RC acts on the new rows introduced by RI, so RC will always follow RI. Consequently, (RI, RD, RC), (RI, RC, RD) and (RD, RI, RC) are the only possible permutations. We note the following observations regarding \sys{}.

\begin{itemize}
    \item \textit{Row insertion assists the performance of row deletion and row completion.} Our reward functions are based on MEM likelihood of the entities occurring in the dialog (eq. \ref{eq:mem_fn}). When an entity (say a restaurant) in a dialog is missing from the KB, eq \ref{eq:mem_fn} yields a very low likelihood value. Consequently, training of RD and RC is adversely affected as reward functions become uninformative on such dialogs. By ensuring that training dialogs do not contain entities missing from the training KB, RI assists the training of RD and RC.
    \item \textit{RD assists training of RC.} Among row deletion and completion, RL training of RC is challenging due to larger action space. We thus run RD first to remove rows from the KB that disrupts the reasoning in the dialogs. This further helps RC during training.
\end{itemize}


We experiment with the these three orderings on inc-bAbI(M) dataset and report the results in table 4. (RI, RD, RC) outperforms the other two permutations as expected. We note that dropping RI leads training dialogs to contain entities missing from the KB. Further, it adversely affects the training of other DKAF models. Similarly, dropping RD leaves training KB with rows that upset dialog reasoning patterns and also disrupt RC training. Finally, dropping RC does not influence the preceding models. As a result, we expect dropping RI should cause a higher drop in performance followed by RD and RC as discussed in section \ref{exp:ablation}.

\begin{table}[t]
\centering
\small
\begin{tabular}{lcc}
\toprule
 \textbf{Permutation}& \textbf{Dialog Acc.} & \textbf{Response Acc.} \\ \midrule
\textbf{(RI, RD, RC)}        & \textbf{88.6}                             & \textbf{99.70}                           \\
(RD, RI, RC)        & 83.6                             & 99.01                           \\
(RI, RC, RD)        & 86.8                             & 99.17                         \\ \midrule

\end{tabular}
\caption{Different orderings of models in \sys{}.}
\label{tab:DKAF_order}
\end{table}

\subsection{\sys{} Model Evaluations}
We evaluate RI, RD, and RC models for their corresponding tasks. Table \ref{tab:dkaf_models} summarizes our findings. For a given dialog $d$, we identify set $R$ of rows by comparing training KB $\mathcal{K}_T$ with contemporary KB $\mathcal{K}_d$ for the dialog. We then use $R$ to compute F1 for RI. We observe that RI model performs reasonably well in both \incbabi{} and \incBiTOD{} datasets though we observe a performance drop in case \incBiTOD{}. This is expected as \incBiTOD{} is human-generated and provides a more challenging setting.

For RD model, we obtain set $D_g$ of rows that occur in $\mathcal{K}_T$ but are missing from $\mathcal{K}_d$. We compare rows $D_p$ deleted by RD with $D_g$ to compute row deletion F1. We find that performance of RD model is comparatively poor on both the datasets. RD task is difficult compared to RI due to lack of supervision. Further, RD requires understanding of complex reasoning patterns in the datasets. Our RL-based approach alleviates these challenges though there still remains margin for improvement. Nonetheless, we obtain significant performance gains with RD as discussed in \ref{exp:ablation}.

We evaluate RC model on \incbabi{} dataset. In this case, we consider a prediction by the model to be correct if the predicted rating fits into the rating order in the KB. We then report accuracy across all predictions of the RC model.

\begin{table}[]
\centering
\small
\begin{tabular}{llll}
\toprule
Dataset  & RI F1        & RD F1 & RC Acc \\
\midrule
incbAbI  & 1.0 (1.0)    & 0.451 & 0.795  \\
incBiTOD & 0.708 (0.96) & 0.398 &        \\
\bottomrule
\end{tabular}
\caption{\sys{} model evaluation. F1 scores for relationship extraction are given in brackets.}\label{tab:dkaf_models}
\end{table}

\section{Conclusions}
We define the novel task of end-to-end training of task-oriented dialog agents, when training data may have inconsistencies between dialog and accompanying KB. This scenario arises, when KB evolves over time, but only one final KB is attached with the data, instead of saving KB snapshots associated with each training dialog. We also contribute two datasets, curated by systematically modifying bAbI and BiTOD datasets, for our task. 

Existing state-of-the-art TOD models, when trained on our datasets, can get quite confused. Our proposed solution, \sys, hypothesizes corrections to KB for each dialog so that the KB becomes dialog-consistent. Since no explicit annotation is available, the modules for KB correction are trained via distant supervision and reinforcement learning. When trained on such corrected data, \sys-based TOD models outperform vanilla TOD models in almost all settings. We release our code and data for further research on the topic. 

\section*{Acknowledgements}
This work is supported by IBM AI Horizons Network grant, grants by Google, Verisk, and 1MG, an IBM SUR award, and the Jai Gupta chair fellowship by IIT Delhi. Vishal is supported by a Google Fellowship. We also thank the IIT Delhi HPC facility for its computational resources.

\section*{Limitations}
\sys{} model has only been tested on English data so far.  At the moment, we curate new datasets by systematic modification of existing datasets. Our simulation strategy is limited as it does not capture real-world factors (e.g. COVID-19 pandemic) that have a drastic impact on restaurant availability. Finally, It would be interesting to find a real-world dataset and verify whether the proposed methods give similar performance gains on it or not.

\bibliographystyle{acl_natbib}
\bibliography{custom}

\appendix

\section{Dataset Details}
\label{sec:data-stats}
Here we provide details for \incbabi{} and \incBiTOD{} datasets. Table \ref{tab:datasets} shows the train, validation and test splits of the \incBiTOD{} and \incbabi{}. 

\incbabi{} consists of dialogs from restaurant domain where queries the agent for restaurants fitting user constraints. Agent gathers all user constraints and suggests fitting restaurants in descending order. User can further request for address or phone number for the restaurant of their choosing. The restaurant knowledge base consists of 1200 entries where each entry has 8 associated attributes. \incbabi{} dataset has with 35.8\% inconsistent dialogs.

\incBiTOD{} is a multi-domain dataset containing dialogs from hotel, restaurant and attraction domains. In \incBiTOD{}, the agent suggests user (hotel, restaurant or attraction) based on user-provided constraints. There are 699 hotels, 1218 restaurants, and 305 attractions. A hotel, a restaurant, and an attraction have 9, 9, and 6 attributes respectively. \incBiTOD{} dataset has 23\% inconsistent dialogs. Note that we do not simulate attraction KB as they rarely change. We simulate availability of hotels using a Bernoulli process.

\begin{table}[h]
\centering
\small
\resizebox{0.46\textwidth}{!}{
\begin{tabular}{lcccc}
\toprule
\multirow{2}{*}{} & \multirow{2}{*}{\incbabi{}} & \multicolumn{3}{c}{\incBiTOD{}} \\
\cmidrule{3-5}
                  &                          & Hotel      & Restaurant     & Attraction     \\
\midrule
Train Dialogs     & 1000                     & 865        & 465            & 283            \\
Val Dialogs       & 1000                     & 84         & 56             & 29             \\
Test Dialogs      & 1000                     & 142        & 64             & 45             \\
\bottomrule
\end{tabular}}
\caption{No. of dialogs in train, validation and test sets.}
\label{tab:datasets}
\end{table}

\section{\sys{} Details}\label{app:mod_comp}
\sys{} consists of four models - RI, RD, RC, and reward function. We first present component modules present in \sys{} models followed by separate discussion on each model. Finally, we provide training details for \sys{}.

\subsection{Component Modules}
\subsubsection*{Dialog Encoder}\label{app:dialog_encoder}
We use a hierarchical dialog encoder \citep{Sordoni2015AHR} in all the \sys{} models. Our design follows hierarchical attention mechanism from \citep{Yang2016HierarchicalAN}. Hierarchical dialog encoder consists of two components - utterance level encoder and dialog level encoder.

Let $d = [u^{u}_1, u^{a}_1, u^{u}_2, u^{a}_2, ..., u^{u}_m, u^{a}_m]  = [u_1, u_2, ..., u_{2m-1}, u_{2m}]$ be a given dialog with $m$ turns where $u_{i}$ is $i^{th}$ utterance in the dialog.
Let $u_{i} = [w_{i1}, w_{i2},..., w_{i{l_i}}]$ where $w_{ik}$ is encoding for $k^{th}$ token in $u_i$ and $l_{i}$ is number of tokens in $u_i$. Each token is encoded as sum of its token embedding (initialised randomly) and token tag embedding. Here, token tag is the entity type if token is an entity, null otherwise.

Utterance level encoder computes feature vectors for each token in $u_i$ as 
\begin{equation*}
    [h_{i1}, h_{i2},...,h_{i{l_i}}] = BiGRU([w_{i1}, w_{i2},..., w_{i{l_i}}])
\end{equation*}
Encoding $\bm{h}_i$ for each utterance is then computed using Luong attention \cite{Luong2015EffectiveAT} as 
\begin{align*}
    \bm{h}_i &= \sum_{k=1}^{l_i} \alpha_k h_{ik} \\
    \alpha_{k} &= softmax(g_{u}(h_{ik}))
\end{align*}
where $g_{u}(h_{ik})$ is a feed-forward network. Dialog level encoder takes $[\bm{h}_1, \bm{h}_2,...,\bm{h}_{2m}]$ as input and computes dialog feature vector $\bm{c}$ using Luong attention as
\begin{align*}
    [H_1, H_2,...,H_{2m}] &= GRU([\bm{h}_1, \bm{h}_2,...,\bm{h}_{2m}]) \\
    \bm{c} &= \sum_{i=1}^{2m} \beta_i H_i \\
    \beta_i &= softmax(g_{d}(H_i))
\end{align*}
where $g_{d}$ is another feed forward network. Note that the hierarchical dialog encoder outputs hidden vectors for each token in an utterance, each utterance, and the entire dialog.

\subsubsection*{KB Encoder}\label{app:kb_encoder}
KB encoder treats input KB as a relational graph $G=(\mathcal{V}, \mathcal{E}, \mathcal{R})$ where $\mathcal{V}$ and $\mathcal{E}$ are set entities and relationships in KB respectively. $\mathcal{R}$ denotes a set of all relation types based on domain. KB encoder uses $L$-relation graph convolution (r-GCN) layers \citep{Schlichtkrull2018ModelingRD} for computing the KB entity feature. It forms a set $Z_0 = \{z^0_e\}_{\forall e \in \mathcal{V}}$ of entity embeddings as input to the first r-GCN layer. $l^{th}$ GCN layer updates the features for entity $e\in\mathcal{V}$ as
\begin{equation*}
    z^l_e = \sigma\left( \sum_{r \in \mathcal{R}} \sum_{e' \in \mathcal{N}^r_e} W^{(l)}_r z^{(l-1)}_{e'} + W^{(l)}_0 z^{(l-1)}_{e} \right)
\end{equation*}
where $\mathcal{N}^r_{e}$ is set of entities that are related to $e$ in $G$ via relationship type $r$. Matrices $W^{(l)}$s are parameters of the r-GCN layer and $\sigma$ is ReLU activation function. We use $\bm{Z} = \{\bm{z}_e\}_{\forall e \in \mathcal{V}}$ to denote the output of the last ($L^{th}$) r-GCN layer.

\subsubsection*{Memory Network}\label{app:memory_network}
Memory network performs $k$-hop reasoning \citep{Sukhbaatar2015EndToEndMN} over a memory using given input query $q^0$. In our case, KB entity features $\bm{Z}$ form the memory while query $q^0$ depends upon the model (RD, RC or MEM reward model). At $l^{th}$ hop, the memory network refines the query vector using Luong attention as
\begin{align*}
    o^{(l)} &= \sum_{k=1}^{|Z|} \gamma_k \bm{z}_k \\
    \gamma_k &= softmax(g^{l}(\bm{z}_k || q^{(l-1)})) \\
    q^{(l)} &= q^{(l-1)} + o^{(l)}
\end{align*}
where $g^{l}$ is a feed-forward network at $l^{th}$ hop and $||$ is concatenation operator. The output of the memory network is final query vector $\bm{q} = q^{(k)}$.

\subsection{Model Architectures}\label{app:mod_arch}
\subsubsection*{Row Insertion (RI)}\label{app:ri_arch}
For a given input $(d, e_1, e_2, r)$, RI model infuses position indicators for entities $e_1$ and $e_2$ in $d$ as in \citet{Zhang2015RelationCV}. It then encodes utterances in the resulting dialog with utterance level encoder described in section \ref{app:dialog_encoder}. For an utterance $u_i$ in the dialog, RI model appends $\bm{h_i}$ with position vectors $pos_{i_1}$ and $pos_{i_2}$ relative to utterances containing $e_1$ and $e_2$ respectively. The concatenated vector is then passed to the dialog level encoder which computes the dialog feature vector $\bm{c}$.

RI model concatenates dialog features $\bm{c}$ and entity features $h_{e_1}$ and $h_{e_2}$ from the dialog encoder and feeds them to a classification layer for relation type $r$.

\subsubsection*{Row Deletion (RD)}\label{app:rd_arch}
For a given input $(d, K, \rho)$, RD model computes dialog features and KB features using dialog encoder and KB encoder respectively. It computes encoding for the input $\rho$ as $\bm{z}_{\rho} = \sum_{e\in\rho} \bm{z}_{e}$. Finally, it sets initial query $q^0 = \bm{c}$ and reasons over KB entity encoding using memory network to get refined query vector $\bm{q}$. Finally, it concatenates vectors $\bm{q}, \bm{z}_{\rho}$ and passes the resultant through a binary classifier layer.

\subsubsection*{Row Completion (RC)}\label{app:rc_arch}
Let $(d, e_s, r, K)$ be input to RC model. RC model infuses position indicators and position vectors with respect to $e_s$ and encodes resulting dialog using dialog encoder. It encodes $K$ using KB encoder. It forms initial vector $q^0 = f(\bm{c}||h_{e_s})$ where $f$ is a feed-forward layer as input to memory network. Finally, it combines memory network output $\bm{q}$ with entity features $\bm{z}_{e_s}$ and feeds the resultant to a feed-forward layer that performs predictions over $E_r$ of possible target entities.

\subsubsection*{Masked Entity Model (MEM)}\label{app:mem_arch}
Recent works \cite{Wu2019GlobaltolocalMP, He2020Fg2seqEE, Raghu2021ConstraintBK, He2020TaskOrientedDG} use pointer networks that copy entities required in the agent response from dialog history tokens and KB entities. Consequently, we design our MEM model $P(e|H_e, K)$ as a dual pointer network as
\begin{multline*}
    P(e|H_e, K) \\
        = \lambda P_{kb}(e|H_e, K) + (1 - \lambda) P_{ctx}(e|H_e, K)
\end{multline*}
Here $P_{kb}$ and $P_{ctx}$ compute probabilities for copying entity $e$ from KB entities and tokens from masked dialog history $H_e$ respectively. $\lambda$ is a soft-gate to select entity $e$ from $H_e$ and the KB.

MEM model consists of hierarchical dialog encoder, KB encoder and memory network discussed earlier. For a given input $(H_e, K)$, MEM model uses position indicators and features with respect to \textit{<mask>} token and computes dialog features using dialog encoder. It encodes $K$ using KB encoder. It forms initial query $q^0$ to memory network as concatenation dialog features $\bm{c}$ and \textit{<mask>} token features $h_{m}$. It receives $\bm{q}$ as output of the memory network.

MEM model computes $P_{kb}$ over KB entities using Luong attention between concatenated vector $(\bm{q}||h_{m})$ and KB entity encoding $\bm{Z}$. Similarly, it computes $P_{ctx}$ using Luong attention between $(\bm{q}||h_{m})$ and $H_e$ token encoding from dialog encoder. Finally, it computes soft-gate $\lambda = g_2(\bm{q})$ where $g_2$ is a feed-forward network. 

\subsection{Training Details}
We find that following hyper-parameter setting works decently across all \sys{} models. We use input embedding size of 100, learning rate of $1e-4$ and batch size of 32. For RD, RC and MEM models, we use entity embedding size of 100 and 8 r-GCN layers in KB encoder and 8 hops reasoning in the memory network. We train RI, RD, RC and MEM models for 30, 200, 200 and 100 epochs. It takes around 4 hours to train \sys{} for both \incbabi{} and \incBiTOD{} datasets.

Since the problem assumes no annotated data, we use either distant supervision or reinforcement learning to train the models. We track the training progress of each model in \sys{} as follows.

\vspace{1ex}
\noindent \textbf{Row Insertion}
The RI model is relation classifier trained using distantly supervised data. We use classifier accuracy as a metric to measure progress during training. The training and validation accuracy of the RI models over epochs on the \incbabi{} dataset is shown in table \ref{tab:RI eval}.

\begin{table}[t]
\centering
\begin{tabular}{|l|l|l|l|l|l|}
\hline
\textbf{Epoch}                        & \textbf{0}                   & \textbf{5}                 & \textbf{10}                & \textbf{15}                & \textbf{20}                \\ \hline
Train Acc. &  0.784 & 1.0 &  1.0 &  1.0 & 1.0 \\ \hline
 Val Acc.   &  0.775 &  1.0 &  1.0 &  1.0 &  1.0 \\ \hline
\end{tabular}
\caption{Progress of training and validation accuracy of RI on \incbabi{}}
\label{tab:RI eval}
\end{table}

\vspace{1ex}
\noindent \textbf{Row Deletion}
We use RL to train the RD model. We report the average reward across epochs for \incbabi{} dataset in table \ref{tab:RD eval}.

\begin{table}[t]
\centering
\resizebox{0.49\textwidth}{!}{
\begin{tabular}{|l|l|l|l|l|l|}
\hline
\textbf{Epoch}                        & \textbf{0}                   & \textbf{10}                 & \textbf{100}                & \textbf{180}                & \textbf{190}                \\ \hline
 Avg. Reward  & -0.590 &	-0.002 &	0.710 &	0.927 &	0.937
 \\ \hline
\end{tabular}}
\caption{Progress of average reward for RD on \incbabi{}}
\label{tab:RD eval}
\end{table}
\vspace{1ex}
\noindent \textbf{Row Completion}
We use RL to train the row completion model as well. Here too, we report the average reward across epochs for \incbabi{} dataset in table \ref{tab:RC eval}:
\begin{table}[t]
\centering
\resizebox{0.49\textwidth}{!}{
\begin{tabular}{|l|l|l|l|l|l|}
\hline
\textbf{Epoch}                        & \textbf{0}                   & \textbf{10}                 & \textbf{100}                & \textbf{180}                & \textbf{190}                \\ \hline
 Avg. Reward  & -0.649&	-0.255	&0.272&	0.674	&0.883

 \\ \hline
\end{tabular}}
\caption{Progress of average reward for RC on \incbabi{}}
\label{tab:RC eval}
\end{table}

\subsection{\sys{} Model Evaluations}\label{app:dkaf_model_metrics}
\vspace{0.5ex}
\noindent\textbf{Row Insertion F1:} We measure efficacy of RI in extracting correct rows from given dialog $d$. Let $K^{ri}$ denote KB obtained post row insertion. Let $R \subseteq R_d$ be the set of rows that participate in $d$. Note that RI can only extract rows from $R$. We compute F1 with following precision and recall $pr = |R \cap (\mathcal{K}^{ri} \setminus \mathcal{K}_T)| / |(\mathcal{K}^{ri} \setminus \mathcal{K}_T)|$ and $re = |R \cap (\mathcal{K}^{ri} \setminus \mathcal{K}_T)| / |R|$. We now report Macro F1 across all the dialogs.\\

\vspace{0.5ex}
\noindent\textbf{Row Deletion F1:} During simulation, we obtain set $D_g$ of rows in $K_T$ that are misaligned with the dialog. Let $D_p$ denote RD's predicted set of rows for deletion. We compute F1 with following precision and recall $pr = |D_p \cap D_g| / |D_p|$ and $re = |D_p \cap D_g| / |D_g|$. We now report Macro F1 across all the dialogs.\\

\vspace{0.5ex}
\noindent\textbf{Row Deletion F1:} Let $K^{rd}$ denote KB obtained post row deletion. Then, $D_p= K^{T} \setminus K^{rd}$ is set of rows deleted by RD and and $D_g = K_T \setminus K_d$ is gold deletion set. We compute F1 with following precision and recall $pr = |D_p \cap D_g| / |D_p|$ and $re = |D_p \cap D_g| / |D_g|$.  Note that our $K_T \ K_d$ can also contain rows that may be neutral to the task (for example, non-participating restaurants in inc-bAbI). Consequently, the recall we get significantly underestimates the actual model performance.\\

\vspace{0.5ex}
\noindent\textbf{Row Completion Accuracy:} In inc-bAbI, the RC model introduces ratings to the newly added rows. Recommendations in inc-bAbI strictly follow the rating order (higher to lower) of the restaurants in KB. Consequently, we consider a prediction by the RC model to be correct if the predicted rating fits into the rating order in the KB. We then report accuracy across all predictions of the RC model.






\section{Rule-based Baseline}\label{app:rule_based}
We propose a  rule-based KB correction framework with the least possible dataset-specific rules that can be applied to any dataset. The rules of the three components of the framework are as below. We use the same notations that are used to explain the different components of \sys{}.

\vspace{1ex}
\noindent \textbf{Row Insertion}
Let $(e_1, r, e_2)$ be a candidate relationship as defined in \ref{sec:ri} where $e_1$ and $e_2$ are entities in input dialog $d$. We use the following rules for deciding whether relation $(e_1, r, e_2)$ to be added to KB.
\begin{enumerate}
    \item If $e_1$ is missing in the KB, insert a new row for $e_1$.
    \item Add relationship $(e_1, r, e_2)$ to the new row if $e_2$ is the closest type-consistent entity to $e_1$ in the dialog.
    \item If $e_2$ is uniquely associated with some entity in KB (for example phone number of a restaurant), do not insert $(e_1, r, e_2)$ to the new row.
\end{enumerate}

\vspace{1ex}
\noindent \textbf{Row Deletion}
We delete a row from the KB if none of the entities unique to that row occur in the dialog.

\vspace{1ex}
\noindent \textbf{Row Completion}
Rules for row completion are highly dataset specific and require considerable domain expertise. Since \incbabi{} is a synthetic dataset, we can derive a reasonable rule for row completion. Here, we add the rating for newly added restaurants such that the order in which restaurants are suggested in the dialog is respected.
\\

Such a rule-based system may not capitalize on fine-grained patterns present in the data for each domain. Note that with detailed domain knowledge, we can design a rule-based approach for row insertion (RI), row deletion (RD), and row completion (RC), which may work for resolving the dialog-KB inconsistencies to a reasonable extent. But such detailed domain-specific knowledge is not always available or may be expensive to collect for every dataset. In contrast, our proposed \sys{} can be trained to solve dialog-KB inconsistency in any dataset without any extra domain information. 

\section{Training baseline models}\label{app:baseline_training}

We adapt SimpleTOD to end-to-end setting and implement it using HuggingFace library\footnote{https://huggingface.co/}. Please refer \ref{app:baseline_training} for more details.

\subsection{SimpleTOD for end-to-end TOD}\label{app:simpletod}
We adapt the input representation given by \citet{HosseiniAsl2020ASL} to end-to-end TOD setting. Our encoding scheme is given in table \ref{tab:stod_input}. Encoded input is then tokenized using GPT2 tokenizer and passed to the model. During training, the model is optimized for log-likelihood of response given context and KB. During inference, model generates a system response provided context and KB using greedy decoding \citep{HosseiniAsl2020ASL}. For SimpleTOD, we performed grid search on four parameters: learning rate, warm ratio, batch-size and number of epoch for both \incbabi{} and \incBiTOD{}. The hyperparameters for best performance are reported in table \ref{tab:SimpleTOD hyperparameter}.

\begin{table}[]
\centering
\begin{tabular}{|l|l|l|l|l|l|}
\hline
          & lr & warmup & bs & epochs \\ \hline
\incbabi{} & 3e-5      & 0.1          & 32         & 4               \\ \hline
\incBiTOD{}{} & 3e-5      & 0.1          & 32         & 10              \\ \hline
\end{tabular}
\caption{Best Hyperparameters for SimpleTOD for \incbabi{} and \incBiTOD{}}
\label{tab:SimpleTOD hyperparameter}
\end{table}

\subsection{GLMP and CDNet}\label{app:CDNet and FG2Seq hyp}
For CDNet and GLMP we are using the same hyper-parameters as mentioned in their respective original papers. The hyperparameters that give us the best results for both \incbabi{} and \incBiTOD{} are mentioned in the table \ref{tab:FG2seq/CDNet hyperparameter}.  For GLMP, we obtain the best performance at one of two values of number of hops mentioned in the table.
\begin{table}[]
\centering
\begin{tabular}{|l|l|l|l|}
\hline
       & learning rate & dropout & no. of hops \\ \hline
GLMP & 1e-4    & 0.1     & 1, 3\\ \hline
CDNet  & 1e-4          & 0.05    & 3            \\ \hline
\end{tabular}
\caption{Best Hyperparameters for GLMP and CDNet for \incbabi{} and \incBiTOD{}}
\label{tab:FG2seq/CDNet hyperparameter}
\end{table}

We use publicly available implementations for FG2Seq\footnote{https://github.com/scoyer/FG2Seq} and CDNet\footnote{https://github.com/dair-iitd/AggNet} baselines.

\section{Compute Resources}\label{app:Compute Respources} 
All experiments were run on a single Nvidia V100 GPU with 32GB of memory. \sys{} has an average runtime of 4 hours on both \incbabi{} and \incBiTOD{}. The compute time for model training for all three models are mentioned in table \ref{tab:compute resources}. For SimpleTOD, \sys{} modified versions of \incbabi{} and \incBiTOD{} take, the same average compute time as the original datasets.  
\begin{table}[t]
\centering
\begin{tabular}{lll}
\hline
\multicolumn{1}{|l|}{}       & \multicolumn{1}{l|}{\incbabi{}}    & \multicolumn{1}{l|}{\incBiTOD{}}   \\ \hline
\multicolumn{1}{|l|}{GLMP} & \multicolumn{1}{l|}{1 hours} & \multicolumn{1}{l|}{0.5 hour}  \\ \hline
\multicolumn{1}{|l|}{CDNet}  & \multicolumn{1}{l|}{9 hours} & \multicolumn{1}{l|}{7 hours} \\ \hline
\multicolumn{1}{|l|}{SimpleTOD     }  & \multicolumn{1}{l|}{4 hours} & \multicolumn{1}{l|}{2.5 hours} \\ \hline               
\end{tabular}
\caption{Average compute time for all the models for \incbabi{} and \incBiTOD{}}
\label{tab:compute resources}
\end{table}
\section{Domain Specific Analysis}\label{app:dsa}
During our experiments, we found that \sys{} exhibits the same trend across the three domains of \incBiTOD{} dataset: hotels, restaurants, and attractions. We have compared the domain-wise results in table \ref{tab:dsa_results}. It can be observed that SimpleTOD is the best baseline on \incBiTOD{} dataset across all three domains. Also, SimpleTOD trained with \sys{} gives us a gain in performance with the best Entity F1 and KB F1 across all domains. In contrast, rule-based KB correction is performing worse than even SimpleTOD, showing that more domain-specific rules are required to obtain better scores.   

\begin{table}[t]
\centering
\resizebox{0.49\textwidth}{!}{
\begin{tabular}{|l|c|c|}
\hline
                      & \multicolumn{1}{l|}{Response Acc.} & \multicolumn{1}{l|}{Dialog Acc.} \\ \hline
CDNet        & 96.33                              & 64.9                             \\ \hline
CDNet + DKAF & 98.34                              & 79.8                             \\ \hline
\end{tabular}
}
\caption{Incremental KB Analysis}
\label{tab:increment_kb}
\end{table}

\begin{table*}[]
\centering
\resizebox{0.99\textwidth}{!}{
\begin{tabular}{lcccccccccc}
\toprule
\multicolumn{1}{c}{\multirow{2}{*}{Model}} &               &                  &                     & \multicolumn{2}{c}{Hotels}             & \multicolumn{2}{c}{Restaurant}         & \multicolumn{2}{c}{Attraction}         \\
\cmidrule(r){2-10}
\multicolumn{1}{c}{}                       & \textbf{Bleu} & \textbf{Ent. F1} & \textbf{KB Ent. F1} & \textbf{Ent. F1} & \textbf{KB Ent. F1} & \textbf{Ent. F1} & \textbf{KB Ent. F1} & \textbf{Ent. F1} & \textbf{KB Ent. F1} \\
\midrule
GLMP                                   & 15.29         & 0.6743             & 0.6326         & 0.6839             & 0.6316                & 0.6640                 & 0.6279                    & 0.6335                 & 0.6502                   \\
CDNet                                      & 19.37         & 0.7717           & 0.7445              & 0.8188          & 0.7975             & 0.6879          & 0.6440             & 0.6788          & 0.6783            \\
SimpleTOD                              & 20.28 & 0.7862    & 0.7566  & 0.8255   & 0.7966      & 0.7118       & 0.6633          & 0.7233       & 0.7488          \\
SimpleTOD + Rule-based                 & 21    & 0.7611    & 0.7733  & 0.7996    & 0.8023       & 0.6890        & 0.7239           & 0.6962        & 0.7236           \\
SimpleTOD + \sys                       & \textbf{{24.91}} & \textbf{0.8187}    & \textbf{0.8330}  & \textbf{0.8402}    & \textbf{0.8616}       & \textbf{0.7915}       & \textbf{0.7677}          & \textbf{0.7400}       & \textbf{0.8232}          \\
SimpleTOD + DKAF w\textbackslash{}o RI & 19.92 & 0.7779    & 0.7488  & 0.8142   & 0.7891        & 0.7200       & 0.6737          & 0.6840       & 0.7034          \\
SimpleTOD + DKAF w\textbackslash{}o RD & 23.48 & 0.7973    & 0.7924  & 0.8264    & 0.8226      & 0.7422        & 0.7185          & 0.7400       & 0.7949          \\
\bottomrule
\end{tabular}}
\caption{Domain Specific results of \incBiTOD{} dataset}
\label{tab:dsa_results}
\end{table*}

\section{Incremental KB Size Analysis}\label{app:kb_analysis}
In this section, we conducted experiments to check the effect of increase in KB size on the efficacy of DKAF. For our experiments, we systematically increased the size of the KB in inc-bAbI dataset by adding new restaurants to the associated training KB. We reported the finding in table \ref{tab:increment_kb} which shows that the is a limited effect on the expected trend. Because of the constrained input sequence length of SimpleTOD we have conducted this experiment on CDNet.

\section{Human Evaluation Details}
\label{app:human_eval}
Our team of annotators consists of two graduate-level students who volunteered for this task. Each of them has completed a course in either Machine Learning or Natural Language Processing, equipping them with the necessary knowledge and expertise. We have great confidence in the quality of their annotations. Additionally, we conducted a thorough review of a selection of randomly chosen annotated samples and found them to be satisfactory. Inter-annotator agreement was $\kappa = 0.31$\cite{cohen1960coefficient} for the relevance score.

\begin{figure*}
    \centering
    \includegraphics[scale=0.73]{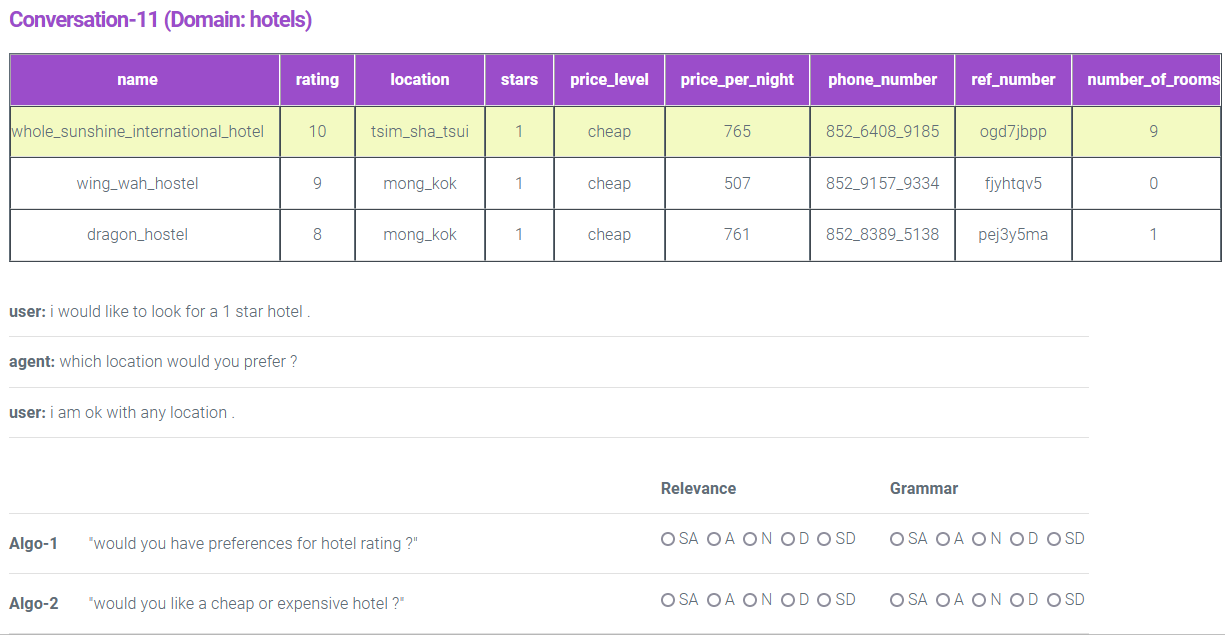}
    \caption{Figure shows a snapshot of the portal used for human evaluation}\label{fig:manual_eval}
\end{figure*}

A snapshot of the portal used for collecting human evaluation is shown in figure \ref{fig:manual_eval}. And the instructions provided to the human annotators are listed below:
\begin{enumerate}
    \item \textbf{What is the task about?}\\
    There are 50 dialog context response pairs in the HTML file. Each context response pair dictates a scenario where user is enquiring the agent about hotels, restaurant,s and attractions to visit. User can optionally request for additional attributes like phone number and address and can make a booking. Agent is expected to suggest hotel, restaurant and attraction with the highest rating among available options.
    Each context response pair has an associated knowledge base (table) where rows corresponding to top-rated entities are highlighted. Along with the context response pair, there are outputs of different dialog systems (randomly shuffled). You are requested to annotate each system-generated output along two dimensions: relevance and grammar, using the following scale:
    \begin{enumerate}
        \item SA: Strongly Agree
        \item A : Agree
        \item N : Neutral
        \item D : Disagree
        \item SD: Strongly Disagree
    \end{enumerate}
    \item \textbf{How to judge relevance?}
    \begin{enumerate}
        \item Strongly Agree - when the generated output conveys the intended information–correct entity (hotel/restaurant/attraction) and its attributes (address, phone, rating, etc). Also, when generated, output requests correct input from the user.
        \item Agree – when generated output contains partial information (e.g., when user request address and phone number but output contains only address).
        \item Neutral – when generated output is hard to decide whether its right or wrong.
        \item Disagree - when the generated response is somewhat unacceptable (e.g., re-querying already known information like cuisine for restaurants and name of the user for booking).
        \item Strongly Disagree – when the generated output contains incorrect information (entities or attributes) for given conversation context.
    \end{enumerate}
    In some cases, generated output contains number of search results of the form \#number. For example, there are $\#3$ available hotels, I recommend \textit{jw\_marriott\_hotel\_hong\_kong} which has a rating of 9.\\
    Since KB provided does not contain this information, you are expected to ignore this term in your evaluation.
    \item \textbf{How to judge grammar?}\\
    The grammar of the response is independent of the dialog context or ground truth. A system output can be marked strongly disagree for relevance and still be marked strongly agree for grammar.
    You can make your own rules about what each rating in the scale means for grammar, but please be consistent with the rules you come up with.
    \item \textbf{Can I use any browser?} \\
    Please use only Firefox as other browsers don't allow you to save the annotations to a json file in your local disk. Before you start the annotation please enter about:config in address bar of Firefox and in the config page set privacy.file\_unique\_origin to False.
    \item \textbf{How do I send you the annotations back?}\\
    After you finish the annotating the file, please click the save annotations button at the bottom of the page. This should save a json file with the same name as the html file in the same folder as the html file. Please send me that json file.

\end{enumerate}

\section{\incbabi{} Examples}\label{app:dkaf_exp}
Table \ref{tab:babi_bitod} demonstrates \sys{} updates to training KB given a dialog context.  Comparison responses generated by SimpleTOD model with and without \sys{} is shown in Table \ref{tab:babi_failed_reasoning}.

\section{\incBiTOD{} Examples}\label{app:bitod_examples}
Table \ref{tab:dkaf_bitod} demonstrates \sys{} updates to training KB given a dialog context. Table \ref{tab:reason_fail} and \ref{tab:hallucination} compares responses generated by SimpleTOD model with and without \sys{}.

\begin{table*}[t]
\small
\resizebox{0.99\textwidth}{!}{
\begin{tabular}{m{0.175\textwidth}m{0.825\textwidth}} 
\toprule
& \textbf{Utterances} \\
\midrule
User & hi \\
Agent & hello what can i help you with today \\
User & can you book a table in \textcolor{blue}{london} \\
Agent & i'm on it \\
User & <SILENCE> \\
Agent & any preference on a type of cuisine \\
User & with \textcolor{blue}{italian} food \\
Agent & how many people would be in your party \\
User & we will be \textcolor{blue}{two} \\
Agent & which price range are looking for \\
User & i am looking for a \textcolor{blue}{expensive} restaurant \\
Agent & ok let me look into some options for you \\
User & <SILENCE> \\
Agent & api\_call \textcolor{blue}{italian} \textcolor{blue}{london} \textcolor{blue}{two} \textcolor{blue}{expensive} \\
User & instead could it be for \textcolor{blue}{four} \\
Agent & sure is there anything else to update \\
User & actually i would prefer in a \textcolor{blue}{moderate} price range \\
Agent & sure is there anything else to update \\
User & actually i would prefer with \textcolor{blue}{british} food \\
Agent & sure is there anything else to update \\
User & no \\
Agent & ok let me look into some options for you \\
User & <SILENCE> \\
Agent & api\_call \textcolor{blue}{british} \textcolor{blue}{london} \textcolor{blue}{four} \textcolor{blue}{moderate} \\
User & <SILENCE> \\
Agent & what do you think of this option: \textcolor{blue}{resto\_london\_moderate\_british\_6stars} \\
User & no i don't like that \\
Agent & sure let me find an other option for you \\
User & <SILENCE> \\
Agent & what do you think of this option: \textcolor{blue}{resto\_london\_moderate\_british\_4stars} \\
User & no this does not work for me \\
Agent & sure let me find an other option for you \\
User & <SILENCE> \\
Agent & what do you think of this option: \textcolor{blue}{resto\_london\_moderate\_british\_1stars} \\
User & let's do it \\
Agent & great let me do the reservation \\
User & what is the phone number of the restaurant \\
Agent & here it is \textcolor{blue}{resto\_london\_moderate\_british\_1stars\_phone} \\
User & do you have its address \\
Agent & here it is \textcolor{blue}{resto\_london\_moderate\_british\_1stars\_address} \\
User & thank you \\
Agent & is there anything i can help you with \\
User & no thanks \\
Agent & you're welcome \\
\bottomrule
\end{tabular}}

\centering
\bigskip
Training KB
\resizebox{0.99\textwidth}{!}{
\begin{tabular}{lccccccc}
\toprule
\textbf{R\_name} & \textbf{R\_phone} & \textbf{R\_cuisine} & \textbf{R\_address} & \textbf{R\_location} & \textbf{R\_number} & \textbf{R\_price} & \textbf{R\_rating} \\
\midrule
resto\_london\_moderate\_british\_4stars & resto\_london\_moderate\_british\_4stars\_phone & british & resto\_london\_moderate\_british\_4stars\_address & london & four & moderate & 4 \\
resto\_london\_moderate\_british\_6stars & resto\_london\_moderate\_british\_6stars\_phone & british & resto\_london\_moderate\_british\_6stars\_address & london & four & moderate & 6 \\
\textcolor{red}{resto\_london\_moderate\_british\_8stars} & \textcolor{red}{resto\_london\_moderate\_british\_8stars\_phone} & \textcolor{red}{british} & \textcolor{red}{resto\_london\_moderate\_british\_8stars\_address} & \textcolor{red}{london} & \textcolor{red}{four} & \textcolor{red}{moderate} & \textcolor{red}{8} \\
\bottomrule
\end{tabular}}

\centering
\bigskip
DKAF updated KB
\resizebox{0.99\textwidth}{!}{
\begin{tabular}{lccccccc}
\toprule
\textbf{R\_name} & \textbf{R\_location} & \textbf{R\_cuisine} & \textbf{R\_number} & \textbf{R\_price} & \textbf{R\_phone} & \textbf{R\_address} & \textbf{R\_rating} \\
\midrule
resto\_london\_moderate\_british\_4stars & london & british & four & moderate & resto\_london\_moderate\_british\_4stars\_phone & resto\_london\_moderate\_british\_4stars\_address & 4 \\
resto\_london\_moderate\_british\_6stars & london & british & four & moderate & resto\_london\_moderate\_british\_6stars\_phone & resto\_london\_moderate\_british\_6stars\_address & 6 \\
\textcolor{green}{resto\_london\_moderate\_british\_1stars} & \textcolor{green}{london} & \textcolor{green}{british} & \textcolor{green}{four} & \textcolor{green}{moderate} & \textcolor{green}{resto\_london\_moderate\_british\_1stars\_phone} & \textcolor{green}{resto\_london\_moderate\_british\_1stars\_address} & \textcolor{green}{2} \\
\bottomrule
\end{tabular}}
\caption{\sys{} updates to training KB given a dialog context. \sys{} adds a new row for \textit{resto\_london\_moderate\_british\_1stars} which appears in the dialog. \sys{} deletes two rows \textit{resto\_london\_moderate\_british\_8stars} that has rating higher than restaurants occurring in dialog upsetting dialog reasoning. \sys{} also adds rating \textit{2} to \textit{resto\_london\_moderate\_british\_1stars} maintaining correct rating order. Insertion and deletions are marked in \textcolor{green}{green} and \textcolor{red}{red} respectively. Entities in dialog are marked in \textcolor{blue}{blue}.}
\label{tab:babi_bitod}
\end{table*}

\begin{table*}
\small
\resizebox{0.99\textwidth}{!}{
\begin{tabular}{m{0.175\textwidth}m{0.825\textwidth}} 
\toprule
& \textbf{Utterances} \\
\midrule
User & hi \\
Agent & hello what can i help you with today \\
User & can you book a table in madrid with indian food for four \\
Agent & i'm on it \\
User & <silence> \\
Agent & which price range are looking for \\
User & i am looking for a moderate restaurant \\
Agent & ok let me look into some options for you \\
User & <silence> \\
Agent & api\_call indian madrid four moderate \\
User & instead could it be with spanish cuisine \\
Agent & sure is there anything else to update \\
User & actually i would prefer in rome \\
Agent & sure is there anything else to update \\
User & no \\
Agent & ok let me look into some options for you \\
User & <silence> \\
Agent & api\_call spanish rome four moderate \\
User & <silence> \\
Agent & what do you think of this option: resto\_rome\_moderate\_spanish\_8stars \\
User & do you have something else \\
Agent & sure let me find an other option for you \\
User & <silence> \\
\bottomrule
\end{tabular}
}

\centering
\bigskip
KB
\small
\resizebox{0.99\textwidth}{!}{
\begin{tabular}{lccccccc}
\toprule
\textbf{r\_name} & \textbf{r\_phone} & \textbf{r\_cuisine} & \textbf{r\_address} & \textbf{r\_location} & \textbf{r\_number} & \textbf{r\_price} & \textbf{r\_rating} \\
\midrule
resto\_rome\_moderate\_spanish\_8stars & resto\_rome\_moderate\_spanish\_8stars\_phone & spanish & resto\_rome\_moderate\_spanish\_8stars\_address & rome & four & moderate & 8 \\
resto\_rome\_moderate\_spanish\_7stars & resto\_rome\_moderate\_spanish\_7stars\_phone & spanish & resto\_rome\_moderate\_spanish\_7stars\_address & rome & four & moderate & 7 \\
resto\_rome\_moderate\_spanish\_6stars & resto\_rome\_moderate\_spanish\_6stars\_phone & spanish & resto\_rome\_moderate\_spanish\_6stars\_address & rome & four & moderate & 6 \\
resto\_rome\_moderate\_spanish\_5stars & resto\_rome\_moderate\_spanish\_5stars\_phone & spanish & resto\_rome\_moderate\_spanish\_5stars\_address & rome & four & moderate & 5 \\
\end{tabular}}

\small
\resizebox{0.99\textwidth}{!}{
\begin{tabular}{m{0.175\textwidth}m{0.825\textwidth}} 
\toprule
\textbf{Model} & \textbf{Response} \\
\midrule
SimpleTOD & what do you think of this option: resto\_rome\_moderate\_spanish\_6stars \\
SimpleTOD + DKAF & what do you think of this option: resto\_rome\_moderate\_spanish\_7stars \\
Gold & what do you think of this option: resto\_rome\_moderate\_spanish\_7stars \\
\bottomrule
\end{tabular}
}
\caption{SimpleTOD incorrectly suggests \textit{resto\_rome\_moderate\_spanish\_6stars} which has lower rating than \textit{resto\_rome\_moderate\_spanish\_7stars}. SimpleTOD + \sys{} suggests correct restaurants.}
\label{tab:babi_failed_reasoning}
\end{table*}

\begin{table*}[t]
\small
\resizebox{0.99\textwidth}{!}{
\begin{tabular}{m{0.175\textwidth}m{0.825\textwidth}} 
\toprule
& \textbf{Utterances} \\
\midrule
User & i am interested in finding a hotel . \\
Agent & how many stars should the hotel you are looking for have , and do you prefer a specific location ? \\
User & the hotel i am looking for should have at least \textcolor{blue}{4} stars and i don't mind where it is located but it's rating should be at least \textcolor{blue}{6} . \\
Agent & alright , do you have any other special requests ? \\
User & that would be all . \\
Agent & i found \#3 hotels from which i would recommend \textcolor{blue}{mira\_moon\_hotel} which has a rating of \textcolor{blue}{9} . \\
User & i want to make a reservation at this hotel . \\
Agent & can you tell me your name please and how many nights you are going to stay at the hotel ? \\
User & my name is \textcolor{blue}{jean} and i will stay there for \textcolor{blue}{1} night . i am in need of having \textcolor{blue}{6} rooms . \\
Agent & can you tell me when you will go there ? \\
User & i will go at the hotel on \textcolor{blue}{july} \textcolor{blue}{5} . \\
Agent & can you please confirm that you want to book \textcolor{blue}{6} rooms for \textcolor{blue}{1} night from \textcolor{blue}{july} \textcolor{blue}{5} at the \textcolor{blue}{mira\_moon\_hotel} hotel ? \\
User & yes , that is what i want . \\
Agent & alright , your reservation has been made and the reference number for it is \textcolor{blue}{3pw8nsim} . \\
\bottomrule
\end{tabular}}

\centering
\bigskip
Training KB
\resizebox{0.99\textwidth}{!}{
\begin{tabular}{lcccccccc}
\toprule
\textbf{name} & \textbf{rating} & \textbf{location} & \textbf{stars} & \textbf{price\_level} & \textbf{price\_per\_night} & \textbf{phone\_number} & \textbf{ref\_number} & \textbf{number\_of\_rooms} \\
\midrule
\textcolor{red}{the\_t\_hotel} & \textcolor{red}{10} & \textcolor{red}{pok\_fu\_lam} & \textcolor{red}{4} & \textcolor{red}{moderate} & \textcolor{red}{806} & \textcolor{red}{852\_0586\_7553} & \textcolor{red}{3v65a5oe} & \textcolor{red}{8} \\
\textcolor{red}{the\_ritzcarlton\_hong\_kong} & \textcolor{red}{10} & \textcolor{red}{kowloon} & \textcolor{red}{5} & \textcolor{red}{expensive} & \textcolor{red}{2134} & \textcolor{red}{852\_6768\_3145} & \textcolor{red}{joaf239b} & \textcolor{red}{4} \\
mier\_serviced\_apartments & 7 & central\_district & 4 & moderate & 885 & 852\_0335\_4038 & rmratcru & 2 \\
\bottomrule
\end{tabular}}

\centering
\bigskip
DKAF updated KB
\resizebox{0.99\textwidth}{!}{
\begin{tabular}{lcccccccc}
\toprule
\textbf{name} & \textbf{rating} & \textbf{location} & \textbf{stars} & \textbf{price\_level} & \textbf{price\_per\_night} & \textbf{phone\_number} & \textbf{ref\_number} & \textbf{number\_of\_rooms} \\
\midrule
\textcolor{green}{mira\_moon\_hotel} & \textcolor{green}{9} &  & \textcolor{green}{4} &  &  &  & \textcolor{green}{3pw8nsim} &  \\
mier\_serviced\_apartments & 7 & central\_district & 4 & moderate & 885 & 852\_0335\_4038 & rmratcru & 2 \\
\bottomrule
\end{tabular}}
\caption{\sys{} updates to training KB given a dialog context. \sys{} adds a new row for \textit{mira\_moon\_hotel} which appears in the dialog. \sys{} deletes two rows \textit{the\_t\_hotel} and \textit{the\_ritzcarlton\_hong\_kong} that have rating higher than \textit{mira\_moon\_hotel} upsetting dialog reasoning. Insertion and deletions are marked in \textcolor{green}{green} and \textcolor{red}{red} respectively. Entities in dialog are marked in \textcolor{blue}{blue}.}
\label{tab:dkaf_bitod}
\end{table*}

\begin{table*}[t]
\resizebox{0.99\textwidth}{!}{
\begin{tabular}{lcccccccc}
\toprule
\textbf{name} & \textbf{rating} & \textbf{location} & \textbf{stars} & \textbf{price\_level} & \textbf{price\_per\_night} & \textbf{phone\_number} & \textbf{ref\_number} & \textbf{number\_of\_rooms} \\
\midrule
alohas\_hostel & 9 & kowloon & 2 & cheap & 614 & 852\_5153\_2856 & tt4jgzs2 & 9 \\
ocean\_inn & 9 & kowloon & 2 & cheap & 632 & 852\_7571\_3796 & 8d2w237w & 10 \\
just\_inn & 8 & tsim\_sha\_tsui & 2 & cheap & 481 & 852\_4281\_9162 & j3i1qob0 & 10 \\
dragon\_hostel & 8 & mong\_kok & 1 & cheap & 761 & 852\_8389\_5138 & pej3y5ma & 1 \\
evergreen\_hotel & 7 & kowloon & 3 & cheap & 699 & 852\_2759\_1053 & hmapwuwm & 2 \\
\end{tabular}}

\small
\resizebox{0.99\textwidth}{!}{
\begin{tabular}{m{0.175\textwidth}m{0.825\textwidth}} 
\toprule
& \textbf{Utterances} \\
\midrule
User & hello , i would need a good hotel to stay . \\
Agent & are you looking for cheap or expensive hotel ? \\
User & i would need a hotel of cheap cost of living . \\
Agent & how about your preferences for the location and the star of the hotel ? \\
User & i am fine with any locations . the hotel would be minimum 1 stars rating . \\
\end{tabular}}

\small
\resizebox{0.99\textwidth}{!}{
\begin{tabular}{m{0.175\textwidth}m{0.825\textwidth}} 
\toprule
\textbf{Model} & \textbf{Response} \\
\midrule
SimpleTOD & there are \#2 hotels available according to your preferences . i would recommend \textcolor{red}{dragon\_hostel} with rating of 8 . \\
SimpleTOD + DKAF & i found \#3 hotels from which i would recommend the \textcolor{green}{alohas\_hostel} which has a rating of 9 . \\
Gold & ok , there are \#5 available hotels that match your requirements . i would recommend \textcolor{green}{alohas\_hostel} with 9 stars rating . \\
\bottomrule
\end{tabular}}
\caption{SimpleTOD incorrectly suggests \textit{dragon\_hostel} which has lower rating than \textit{alohas\_hostel}. SimpleTOD + \sys{} suggests correct hotels. }
\label{tab:reason_fail} 
\end{table*}

\begin{table*}[t]
\resizebox{0.99\textwidth}{!}{
\begin{tabular}{lcccccccc}
\toprule
\textbf{name} & \textbf{rating} & \textbf{location} & \textbf{address} & \textbf{cuisine} & \textbf{dietary\_restrictions} & \textbf{price\_level} & \textbf{phone\_number} & \textbf{ref\_number} \\
\midrule
gaia\_ristorante & 8 & sheung\_wan & gf\_grand\_millennium\_plaza\_181\_queens\_road\_central\_sheung\_wan & italian & gluten\_free\_options & expensive & 852\_2167\_8200 & gpwwc87k \\
\end{tabular}}

\small
\resizebox{0.99\textwidth}{!}{
\begin{tabular}{m{0.175\textwidth}m{0.825\textwidth}} 
\toprule
& \textbf{Utterances} \\
\midrule
User & hello , assistant ? \\
Agent & hello , how may i help you today ? \\
User & i would like a restaurant . \\
Agent & what sort of price and rating would you like ? \\
User & i would like an expensive restaurant with a rating of at least 3 . i don't want any russian cuisine . \\
Agent & what sort of location would you like ? do you have any dietary restrictions ? \\
User & i would like gluten\_free\_options near sheung\_wan . \\
\end{tabular}}

\small
\resizebox{0.99\textwidth}{!}{
\begin{tabular}{m{0.175\textwidth}m{0.825\textwidth}} 
\toprule
\textbf{Model} & \textbf{Response} \\
\midrule
SimpleTOD & there are \#2 restaurants that match your requirements . i would recommend \textcolor{red}{falafel\_tlv} . it serves middle\_eastern food and has a rating of 9 . \\
SimpleTOD + DKAF & i found a restaurant called \textcolor{green}{gaia\_ristorante} . it serves italian food and has a rating of 8 . it is located at gf\_grand\_millennium\_plaza\_181\_queens\_road\_central\_sheung\_wan and the phone number is 852\_2167\_8200 . \\
Gold & i recommend \textcolor{green}{gaia\_ristorante} , which offers italian food and has a rating of 8 . it is located at gf\_grand\_millennium\_plaza\_181\_queens\_road\_central\_sheung\_wan . you can call them at 852\_2167\_8200 . \\
\bottomrule
\end{tabular}}
\caption{SimpleTOD hallucinates \textit{falafel\_tlv} hotels which does not exist in the KB. }
\label{tab:hallucination} 
\end{table*}

\begin{table*}[t]
\resizebox{0.99\textwidth}{!}{
\begin{tabular}{ | m{0.1\textwidth} | m{0.9\textwidth} | } 
  \hline
  Context & \textcolor{blue}{{[}context{]}} \textcolor{cyan}{{[}usr{]}} good morning \textcolor{lime}{{[}sys{]}} hello what can i help you with today ... \textcolor{cyan}{{[}usr{]}} do you have something else \textcolor{blue}{{[}endofcontext{]}} \\ 
  \hline
  KB & \textcolor{red}{{[}db{]}}\textcolor{orange}{{[}row{]}} 60\_west\_hotel location sheung\_wan, phone\_number 852\_9359\_3222, ..., ref\_number 9xkg8w7n \textcolor{orange}{{[}endofrow{]} {[}row{]}} akvo\_hotel location sheung\_wan, phone\_number 852\_6317\_4051, ..., ref\_number ud0jiny2 \textcolor{orange}{{[}endofrow{]}}\textcolor{red}{{[}endofdb{]}} \\
  \hline
  Response & \textcolor{green}{{[}response{]}} i found \#3 hotels from which i would recommend 60\_west\_hotel which has a rating of 8 . \textcolor{green}{{[}endofresponse{]}} \\
  \hline
\end{tabular}}
\caption{SimpleTOD input representation for end-to-end TOD task}
\label{tab:stod_input} 
\end{table*}

\end{document}